\documentclass[10pt,twocolumn,letterpaper]{article}

\usepackage[pagenumbers]{cvpr}      

\usepackage[dvipsnames]{xcolor}

\definecolor{cvprblue}{rgb}{0.21,0.49,0.74}
\usepackage[pagebackref,breaklinks,colorlinks,citecolor=cvprblue]{hyperref}

\usepackage{graphicx}
\usepackage{amsmath}
\usepackage{amssymb}
\usepackage{booktabs}
\usepackage{algorithm}
\usepackage[noend]{algpseudocode}
\algnewcommand{\LineComment}[1]{\State \(\triangleright\) #1}
\usepackage{xcolor}
\usepackage{comment}
\usepackage{siunitx} 
\usepackage{booktabs}
\usepackage{multirow, makecell}
\usepackage{rotating}
\usepackage{subcaption}
\usepackage{siunitx}

\usepackage{comment}

\usepackage{amsmath,amsfonts,bm}

\def\eqref#1{equation~\ref{#1}}

\def\1{\bm{1}}

\def\vx{{\bm{x}}}

\def\vz{{\bm{z}}}

\DeclareMathAlphabet{\mathsfit}{\encodingdefault}{\sfdefault}{m}{sl}
\SetMathAlphabet{\mathsfit}{bold}{\encodingdefault}{\sfdefault}{bx}{n}

\newcommand{\public}{{tar}}
\newcommand{\private}{{res}}
\usepackage{amsmath,amssymb, bm} 
\usepackage{arydshln}
\usepackage{subfiles}
\def \hfillx {\hspace*{-\textwidth} \hfill}

\usepackage{booktabs} 

\usepackage[capitalize]{cleveref}
\crefname{section}{Sec.}{Secs.}
\Crefname{section}{Section}{Sections}
\Crefname{table}{Table}{Tables}
\crefname{table}{Tab.}{Tabs.}

\usepackage{newfloat}
\usepackage{listings}
\floatstyle{ruled}
\newfloat{listing}{tb}{lst}{}
\floatname{listing}{Listing}

\title{Learning Private Representations through Entropy-based Adversarial Training}

\author{Tassilo Klein\\
SAP SE\\
{\tt\small tassilo.klein@sap.com}
\and
 Moin Nabi \thanks{Currently at Apple}\\
SAP SE \\
{\tt\small m.nabi@sap.com}
}

\begin{document}
\algnewcommand\algorithmicinput{\textbf{INPUT: }}
\algnewcommand\Input{\item[\algorithmicinput]}
\algnewcommand\algorithmicoutput{\textbf{OUTPUT: }}
\algnewcommand\Output{\item[\algorithmicoutput]}

\algnewcommand\algorithmicinit{\textbf{INITIALIZATION: }}
\algnewcommand\Init{\item[\algorithmicinit]}
\maketitle
\begin{abstract}
How can we learn a representation with high predictive power while preserving user privacy? We present an adversarial representation learning method for sanitizing sensitive content from the learned representation.
Specifically, we introduce a variant of entropy - \emph{focal entropy}, which mitigates the potential information leakage of the existing entropy-based approaches. 
We showcase feasibility on multiple benchmarks. The results suggest high target utility at moderate privacy leakage. 
\end{abstract}    
\section{Introduction}
\label{sec:introduction}
The intersection of machine learning with privacy and security has emerged as a critical area of research, spurred by the exponential growth of big data and the proliferation of deep learning techniques. While these advancements have led to remarkable progress in diverse fields, they have also amplified concerns about user privacy.  As machine learning-powered services become increasingly integrated into our daily lives, handling sensitive user data necessitates robust privacy-preserving mechanisms to mitigate the risk of privacy creep~\cite{narayanan2006break,backstrom2007wherefore}.

Significant efforts within the machine learning community have focused on developing algorithms that balance preserving user privacy and maintaining acceptable predictive power. Numerous approaches have been proposed, with a prominent strategy being direct data anonymization. This typically involves obfuscating sensitive data components or introducing random noise, effectively controlling privacy and utility trade-offs. For example, data-level Differential Privacy~\cite{dwork2006differential} offers strong theoretical privacy guarantees by injecting calibrated noise into the dataset. However, while theoretically sound, such approaches often incur substantial computational overhead, hindering their practical implementation in complex, state-of-the-art deep learning architectures~\cite{shokri2015privacy}. 

 \begin{figure}
\centering
\includegraphics[width=0.4\textwidth]{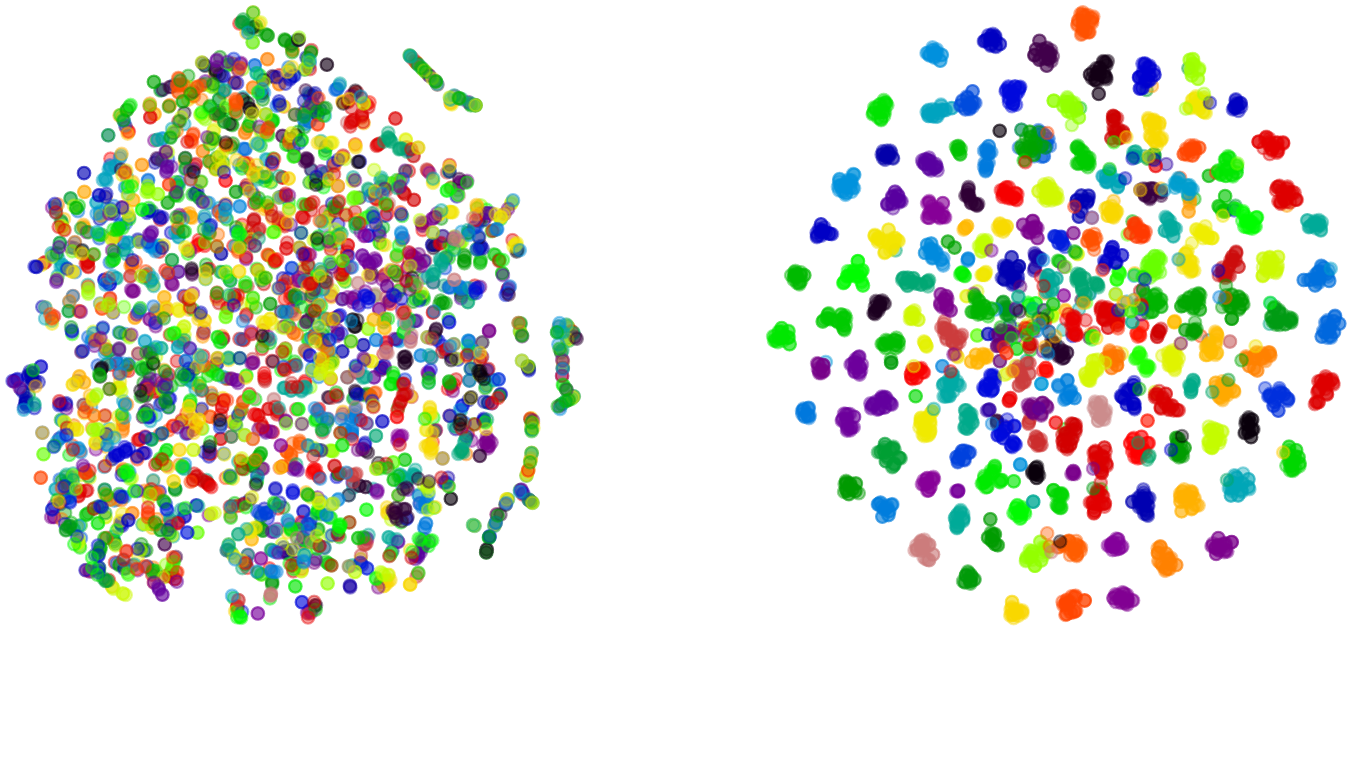}
\caption{t-SNE visualization of the representation extracted from face images. \textbf{Left: } embedding of the private (aka privacy-preserving) representation learned by our method. \textbf{Right:} embedding of privacy-revealing embedding. The absence of clusters in private representation indicates face identities are preserved (each color shows an identity).
}
\label{fig:tsne}
\end{figure}
The pursuit of privacy-preserving machine learning models has led to the exploration of various approaches. Federated learning \cite{mcmahan2016communication,geyer2017differentially,shen2023share,bercea2022federated}, for instance, enables collaborative training on decentralized datasets by exchanging model parameters instead of raw data. While promising, this method grapples with communication overhead and computational complexities. Other strategies involve learning encoded data representations on the client side, but this can still inadvertently leak private information \cite{osia2017privacy,osia2018deep,salem2019updates}, or homomorphic encryption that allows operating on encrypted data~\cite{10581983,Yonetani2017PrivacyPreservingVL}.

Adversarial Representation Learning (ARL) has emerged as a potential solution by leveraging the power of adversarial learning \cite{louizos2015variational,xie2017controllable,madras2018learning} to control the encoding of private information in learned representations \cite{roy2019mitigating,Sadeghi_2019_ICCV}. In ARL, a ``predictor'' aims to extract desired target attributes while an ``adversary'' attempts to infer private attributes from the learned representation. This adversarial game seeks to achieve a representation that retains utility for the target task while minimizing the leakage of sensitive information. 
Two primary strategies exist for training the adversary: (i) ML-ARL, which maximizes the adversary's loss by minimizing the negative log-likelihood of sensitive variables \cite{pittaluga2019learning}, and (ii) MaxEnt-ARL, which maximizes the adversary's entropy, pushing it towards a uniform distribution over sensitive labels \cite{roy2019mitigating,sarhan2020fairness}. However, existing ARL approaches often suffer from an imbalance in learning speeds between the predictor and adversary \cite{NIPS2016_8a3363ab}, leading to suboptimal privacy preservation. We propose a novel framework based on ``focal entropy'' maximization to address this limitation. This method enhances the adversary's task complexity by focusing confusion on classes with high similarity in terms of the sensitive attribute, thereby preventing trivial solutions and promoting more effective privacy sanitization.

Our approach decomposes the latent representation into two components: a  \emph{target embedding} capturing information relevant to the downstream task and a  \emph{residual embedding} absorbing private information. We obtain a privacy-sanitized target embedding suitable for downstream use by discarding the residual embedding after training.  Figure \ref{fig:tsne} showcases t-SNE visualizations of the target and residual embeddings, illustrating the effective segregation of private information. 

\textbf{Contributions:}  \textbf{(i)} the novel application of focal entropy, an off-centered entropy measure, for learning privacy-sanitized representations; and \textbf{(ii)} empirical validation of focal entropy on privacy and fairness benchmarks, demonstrating strong performance outperforming state-of-the-art approaches on several benchmarks.

\begin{figure*}[t!]
\centering
\includegraphics[width=1.0\textwidth]{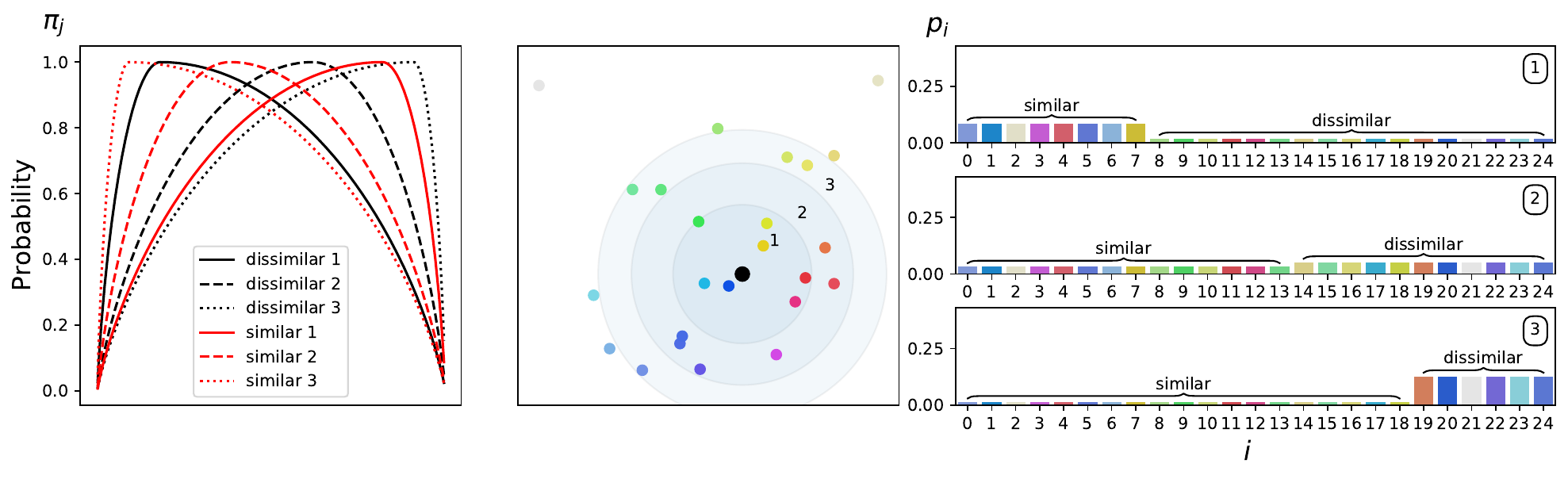}
\caption{Illustration of focal entropy. \textbf{Center:} Visualization of a sample configuration; schematic focus regions depicted as circles ranging from narrow (1) to wide (3). \textbf{Left:} Visualization of probabilities associated with off-centering entropy (similar, dissimilar) for different focus regions scenarios. The more narrow the focus, the more weight ``\emph{similar}'' samples have. The wider the focus range, the more equiprobability is approached. \textbf{Right:} Group-probability visualization for different focus scenarios.}
\label{fig:focalentropy-illustration}
\end{figure*}

\section{Related Works}

The pursuit of privacy in machine learning has led to diverse approaches. Traditional methods often focus on data or parameter-level privacy, such as differential privacy techniques like data anonymization or randomized mechanisms in the learning process  \cite{dwork2017exposed,dwork2006differential,ryoo2017privacy,abadi2016deep}. While our work could incorporate such techniques, for example, during post-classifier training, we differentiate ourselves by aiming to learn inherently private representations. 

The shift towards learning and transmitting representations instead of raw data, though promising  \cite{osia2017privacy,osia2018deep,hamm2017minimax}, introduces new privacy concerns as representations can still leak sensitive information.  Consequently, methods for learning privacy-sanitized representations have emerged, employing techniques like learning representations invariant to sensitive information or disentangling latent features into distinct target and sensitive subsets.

Adversarial learning, in particular, has shown promise for learning privacy-preserving representations across various domains. Examples include learning representations for automatic speech recognition \cite{srivastava2019privacy}, student click-stream data analysis \cite{yang2018learning}, and medical image analysis while obfuscating sensitive information \cite{li2019deepobfuscator,jeong2023noisy,kim2019privacy}. Several works, like ours, utilize adversarial optimization between the encoder and estimators of private attributes to sanitize representations \cite{pittaluga2019learning,xiao2020adversarial,dusmanu2021privacy}.

This paper introduces a novel approach to adversarial sanitization that leverages entropy in a unique way. Our work distinguishes itself by incorporating focal entropy while drawing inspiration from existing methods utilizing entropy and adversarial learning for sanitized representations, such as \cite{roy2019mitigating,Sadeghi_2019_ICCV}. Unlike conventional entropy employed in prior work, focal entropy allows for a more refined approach to adversarial sanitization. Moreover, our method diverges from approaches like \cite{feutry2018learning,gong2020jointly} by eliminating the dependence on downstream task labels during representation learning. Other notable related works exploring representation learning and privacy include  \cite{bouchacourt2018multi,ganin2014unsupervised,meden2021privacy}.

Our work also intersects with the field of disentanglement, which strives to minimize mutual information between feature subsets \cite{10.5555/3327144.3327186}. Various methods have been proposed to achieve this objective. For instance, FFVAE~\cite{creager2019flexibly} leverages total correlation~\cite{10.5555/3327144.3327186} for disentangling sensitive information. Similarly, GVAE~\cite{ding2020guided} employs adversaries to minimize the leakage of private information within the latent code. Recognizing the practical challenges of achieving perfect decorrelation, FADES~\cite{Jang_2024_CVPR} aims to establish conditional independence between target and sensitive information. In contrast, ODVAE~\cite{sarhan2020fairness} pursues fair representation learning without relying on adversarial training, instead enforcing orthogonality in the learned subspace through priors.  Further work in this direction include  \cite{locatello2019fairness,quadrianto2019discovering,10.1145/3534678.3539302,10.1007/s11063-022-10920-8,higgins2017betavae}.  Our work, combines the two major research direction and to the best of our knowledge, is the first to introduce class similarity into the adversary's entropy function, leading to enhanced privacy preservation.

\section{Proposed Method}

We consider the scenario in which we want to learn a feature representation with two objectives. The first objective entails providing high accuracy in terms of target attribute classification. The second objective entails minimizing information leakage w.r.t. sensitive attributes, ideally making the classification of sensitive attributes impossible for an adversary. To this end, we learn an encoder tasked with disentangling the representation by decomposing it into two parts: the target and residual partition. Whereas the (sanitized) target partition allows classification for the target attributes, the residual partition captures the information w.r.t. the sensitive attribute.

\noindent\textbf{Overview:}
The representation learning problem is formulated as a game among multiple players: an encoder \textbf{E}, a target predictor \textbf{T}, a sensitive attribute predictor \textbf{S}, two adversarial classifiers $\mathbf{\tilde{T}}$ and $\mathbf{\tilde{S}}$.
Whereas the non-adversarial predictors enforce the utility and the presence of the associated information in the representation, the adversarial predictors inhibit undesirable information leakage and are directly responsible for the disentanglement and sanitization of the representation. As the adversarial classifiers are learned during training, they act as surrogate adversaries for unknown oracle post-classifiers. 
The setup involves observational input $\vx \in \mathcal{X}$, $M$ \underline{t}arget attributes $\mathbf{t} \in \mathcal{A}_T = A_1 \times A_2 ... A_M$, each having $m_t$ classes $A_t=\{ a_1,...,a_{m_t}\}$, and a \underline{s}ensitive attribute ${s}\in\mathcal{A}_S = \{1, ..., N\}$ of $N$ classes. 
The goal is to yield data representations from an encoder subject to privacy constraints. The representation is learned jointly with several predictors. 
The encoder $
{\vz = q(\vz|\vx;\theta_E): \mathcal{X}\rightarrow\mathcal{Z}}$, parameterized by $\theta_E$ produces an embedding $\vz$ in the latent space $\mathcal{Z}$. For the sake of privacy, we employ a modified encoder.  
We propose an encoder that splits the generated representation into two parts: $\vz_\public \in \mathcal{Z}_1$ (target part), $\vz \in \mathcal{Z}_2$ (residual part). 
Without loss of generality, we assume that both parts are of equal dimensionality. Encoding of the representation entails a stack of shared initial layers that diverge into parallel stacks from the penultimate layer onward into two \emph{separate} network streams to disentangle the information.
To this end, we let $
{\vz_\public = q_\public(\vz_\public|\vx;\mathbf{\theta}_E^\public): \mathcal{X}\rightarrow\mathcal{Z}_1}$ denote the encoder of the target representation stream and the encoder of the $
{\vz_\private = q_\private(\vz_\private|\vx;\mathbf{\theta}_E^\private) : \mathcal{X}\rightarrow\mathcal{Z}_2}$ the residual stream, respectively.  
Next, we define the predictors for the target and sensitive attributes. The target predictor is given as $
{p_T(\mathbf{t}|\vz_{\public};\theta_{\public}): \mathcal{Z}_1\rightarrow\mathcal{A}_T}$. The sensitive attribute predictors are given as $
{p_S(s|\vz_{\private};\theta_\private): \mathcal{Z}_2\rightarrow\mathcal{A}_S}$, parameterized by $\theta_\public$ and $\theta_\private$. Last, the associated adversarial predictors are denoted as $
{\tilde{p}_T(\mathbf{t}|\vz_\private;\tilde\theta_\private): \mathcal{Z}_2\rightarrow\mathcal{A}_T}$ and $
{\tilde{p}_S(s|\vz_\public;\tilde\theta_\public): \mathcal{Z}_1\rightarrow\mathcal{A}_S}$, parameterized by $\tilde\theta_\public$ and $\tilde\theta_\private$. Note that the difference between predictors and their adversary is their \emph{swapped} input source.

\noindent\textbf{Optimization:}
Learning the representation formalizes as the optimization of a multi-player nonzero-sum game:
\begin{equation}
\begin{split}
\min_{\mathbf{\theta}_{E,\private,\public}} \max_{\tilde\theta_{\private,\public}}  \begingroup\color{NavyBlue}\alpha_T\phi_T(\mathbf{\theta}_E,\theta_{\public})\endgroup + \\ \begingroup\color{NavyBlue}\alpha_S\phi_{S}(\mathbf{\theta}_E,\theta_{\private})\endgroup +  \begingroup\color{OrangeRed}\beta_{\tilde{T}}\phi_{\tilde{T}}(\mathbf{\theta}_E,\tilde\theta_\private)\endgroup + \begingroup\color{OrangeRed}\beta_{\tilde{s}}\phi_{\tilde{S}}(\mathbf{\theta}_E,\tilde\theta_\public)\endgroup
\label{eq:cost_function}
\end{split}
\end{equation}

with $\phi_*(.|.)$ denoting the players and $\lambda, \alpha_*, \beta_*\in\mathbb{R}$ weighting scalars.
The hyperparameters allow for a trade-off between utility and the latent code's privacy preservation. 
As predictors and adversarial predictors have a competitive relationship, they are optimized differently. Predictors are trained using cross-entropy minimization, whereas adversarial predictors maximize entropy. 

\subsection{Entropy-based Adversarial Training}

\noindent\textbf{Predictors:}
The predictors are all modeled as conditional distributions and trained by cross-entropy minimization:
\begin{align}
\begingroup\color{NavyBlue}
\phi_{T}(\mathbf{\theta}_E, \theta_\public) = D_{KL}(p(\mathbf{t}|\vx)\| {p}_{T}(\mathbf{t}|\vz_\public);\theta_\public) \endgroup \\
\begingroup\color{NavyBlue}
\phi_{S}(\mathbf{\theta}_E,\theta_\private) = D_{KL}(p(s|\vx)\|{p}_S(s|\vz_\private);\theta_\private) \endgroup
\end{align}

Here $p(\mathbf{t}|\vx)$ and $p(s|\vx)$ denote the ground truth labels for the training input $x$ for the target attribute and the sensitive attribute, respectively. $D_{KL}(.\|.)$ denotes the Kullback-Leibler divergence.

\noindent\textbf{Sanitization:}
To minimize the information leakage across representation partitions, we leverage entropy maximization. This enforces the representation to be maximally ignorant w.r.t. attributes in the sense of domain confusion. Specifically, for the adversarial target attribute, we let
\begin{equation}
\begingroup\color{OrangeRed}
 \phi_{\tilde{T}}(\mathbf{\theta}_E,\tilde\theta_\private) = D_{KL}(\tilde{p}_T(\mathbf{t}|\vz_\private)\|\mathcal{U};\tilde\theta_\private)
 \endgroup
 \end{equation}

where $\mathcal{U}$ denotes the uniform distribution. Although maximizing entropy is sufficient for non-private attributes to minimize information leakage across representation partitions, we postulate that proper sanitization must be conducted w.r.t. focus classes in a similarity-aware fashion. We aim to sanitize the information of each observation within a focus area, that is, the nearest neighbors (NN). For each observation, the NNs share the most commonalities; hence, separation w.r.t. NNs is less likely to be trivially achievable. 

\subsection{Focal Entropy}

Our method requires partitioning the sensitive attribute into two sets w.r.t. mutual discriminativeness: ``\emph{\textbf{s}imilar}'' and ``\emph{\textbf{d}issimilar}''. We let $N_{s},  N_{d}$ denote the number of classes in the respective set, with $N_{s} \ll N_{d}$ typically holding true, and $N=N_{s}+N_{d}$. 
The partitioning in similar and dissimilar is input specific and is conducted either according to: \textbf{a)} label information, or \textbf{b)} using some scoring function obtained using a pre-trained model or ``\emph{on-the-fly}'' during training. Thus, let $\mathbf{r}(\vx)\in \mathbb{N}^N$ be the scores given the observational input $\vx$, which we will simply denote by $\mathbf{r}$ for the sake of the economy of notation. Then group members associated with ``\emph{similar}'' consist of the set of labels corresponding to the $k$-largest scores, and  ``\emph{dissimilar}'' the complement:
\begin{equation}
\mathcal{A}_{similar}= \left\{ s_i \in \mathcal{A}_S^{(k)}:\forall i \in \{1,...,k\}, r_{s_i} \geq r_{[k]}\right\}
\end{equation}

Here, $r_{[k]}\in \mathbb{N}$ denotes the $k$-th largest element of $\mathbf{r}$, and $\mathcal{A}_S^{(k)}$ denotes the set of $k$ distinct elements in $\mathcal{A}_S$ with $\mathcal{A}_{dissimilar}=  \mathcal{A}_S \setminus \mathcal{A}_{similar}$. Given the just defined notion of similarity, focal entropy aims at establishing uniformity in terms of likelihood (w.r.t. the sensitive attribute) within each group -- with the probability mass divided \emph{equally} between the two groups (similar) and (dissimilar). Given $N_{s} \neq N_{d}$, this implies the re-weighting of the classes. Assuming $N_{s} \ll N_{d}$, this aims at to give proportionally more weight to confusion with respect to members of similar classes. Analogously, members of dissimilar classes are down-weighted accordingly. Consequently, the classifier is forced to be maximally ignorant w.r.t. the class properties of the similar class members that share a high correlation. The feature representation produced by the encoder should not have sensitive information in the target part. This approach implicitly drives sanitization by focusing on specific targets, so we refer to this as ``\emph{focal entropy}''. 
Implementation of the focal entropy criterion is equivalent to maximization w.r.t. an off-centered entropy~\cite{lallich2007} in the special case of normalized uniform probability within each group.
That is in contrast to conventional entropy, which takes its maximal
value when the distribution of the class variable is uniform $(\frac{1}{N},...,\frac{1}{N})=\mathcal{U}$, focal-entropy takes its maximum off-centered. 
See Fig.~\ref{fig:focalentropy-illustration} for the schematic illustration of the proposed concept. Specifically, focal entropy seeks to have the entropy peak at $\bm{\tau}\in \mathbb{R}^N$, with $\bm{\tau} \neq \mathcal{U}$. Assuming equally divided probability mass, we yield $\bm{\tau}$ defined as:

\begin{equation}
\begin{split}
\tau_{1,..,N_{s}}=\frac{N_{d}}{N_{s}^2+N_{d} N_{s}}, \tau_{N_{s}+1,..,N}= \frac{N_{s}}{N_{d}^2+N_{s} N_{d}}
\end{split}
\end{equation}

wither $\sum_i^N\tau_i=1$. Further, let $h(.)$ denote the entropy w.r.t. a vector $\mathbf{p}=(p_1,...,p_N) \in \mathbb{R}^N$ of probabilities with $\sum_i^N p_i=1$ defined as $h(\mathbf{p}) = \sum_{j=1}^N p_j \log p_j$. In order to make the entropy assume its maximum at $\bm{\tau}$, the class probabilities $\mathbf{p}$ have to be transformed. To that end, each $p_i \in \left[0,1\right]$ is mapped to a corresponding $\pi_j$, according to:
 \begin{equation}
 \begin{split}
     \pi_j = \frac{p_j}{N\tau_j} \quad\text{ if } 0 \leq p_j \leq \tau_j,\\ \pi_j = \frac{N(p_j-\tau_j)+1-p_j}{N(1-\tau_j)}\quad\text{ if } \tau_j \leq p_j \leq 1.
\end{split}
 \end{equation}

\begin{table}
\centering
 \resizebox{8.5cm}{!}{%
\begin{tabular}{l|ll}
\hline
\multicolumn{3}{c}{\textbf{CIFAR-100}~\cite{Krizhevsky09learningmultiple}} \\
\hline
Method & Tar. Acc. ($\uparrow$) & Adv. Acc. ($\downarrow$)\\
\hline
$\uparrow$ Upper bound & 1.0 & 0.01 \\
\hdashline
ML-ARL~\cite{xie2017controllable} &  0.74 & 0.80 \\
MaxEntARL~\cite{roy2019mitigating}  & 0.71 & 0.16 \\
Ker.-SARL~\cite{Sadeghi_2019_ICCV} & 0.80 & 0.16 \\
ODR~\cite{sarhan2020fairness}  & 0.71 & \textbf{0.14} \\
\hline
\textbf{Ours (\texttt{Focal entropy})}  & \textbf{0.82} & {0.16} \\
\hline
\multicolumn{3}{c}{\textbf{CelebA}~\cite{guo2016ms}} \\
 \hline
Method & Tar. Acc. ($\uparrow$) & Adv. Acc. ($\downarrow$)\\
\hline
$\uparrow$ Upper bound & 1.0 & $<$ 0.001   \\
\hdashline
Szabo~\cite{2018challenges} & -  & 0.09 \\
Cycle-VAE~\cite{Jha_2018_ECCV}&  - & 0.14 \\
DrNet~\cite{DrNet2017}  & - & 0.03 \\
LORD~\cite{Gabbay2020Demystifying} &  - &  \textbf{$<$ 0.01} \\
ML-VAE~\cite{kingma2013autoencoding} & 0.88 & 0.178 \\
Deep-Face-Att. & 0.873 & - \\
PANDA~\cite{oldfield2023panda} & 0.854 & - \\
$D^2AE$~\cite{Liu_2018_CVPR} & 0.878 & - \\
\hline
\textbf{Ours (\texttt{Focal entropy})}  & \textbf{0.90} & \textbf{$<$ 0.01}
\\
\hline
\end{tabular}
}
\caption{\textbf{Privacy sanitization:} Results apart from ours originate from \cite{roy2019mitigating,Sadeghi_2019_ICCV,sarhan2020fairness} and the respective papers.}
\label{tab:results_privacy}
\end{table}

\begin{table}
\centering
 \resizebox{8.5cm}{!}{%
\begin{tabular}{l|ll}
\hline
\multicolumn{3}{c}{\textbf{German}~\cite{Dua:2019}} \\
\hline
Method & Tar. Acc. ($\uparrow$) & Adv. Acc. ($\downarrow$)\\
\hline
LFR & 0.72 & 0.80 \\
ML-ARL~\cite{xie2017controllable} &  0.74 & 0.80 \\
ML-VAE~\cite{kingma2013autoencoding} & 0.72 & 0.79 \\
VFAE~\cite{louizos2017variationalfairautoencoder} & 0.73 & 0.80 \\
MaxEntARL~\cite{roy2019mitigating} & 0.72 & 0.80 \\
Ker.-SARL~\cite{Sadeghi_2019_ICCV} & 0.76 & 0.81 \\
ODR~\cite{sarhan2020fairness} & 0.77 & 0.71 \\
\hline
\textbf{Ours (\texttt{Focal entropy})}  & \textbf{0.80} & \textbf{0.70} \\
\hline
\multicolumn{3}{c}
{\textbf{Adult}~\cite{Dua:2019}} \\
\hline 
Method & Tar. Acc. ($\uparrow$) & Adv. Acc. ($\downarrow$)\\
\hline
LFR & 0.82 & 0.67 \\
ML-ARL~\cite{xie2017controllable} &  0.84 & 0.68 \\
ML-VAE~\cite{kingma2013autoencoding} & 0.82 & 0.66 \\
VFAE~\cite{louizos2017variationalfairautoencoder} & 0.81 & 0.67 \\
MaxEntARL~\cite{roy2019mitigating}  & 0.85 & 0.65 \\
Ker.-SARL~\cite{Sadeghi_2019_ICCV} & 0.84 & 0.67 \\
ODR~\cite{sarhan2020fairness}  & \textbf{0.85} & {0.68} \\
\hline
\textbf{Ours (\texttt{Focal entropy})}  & \textbf{0.85} & \textbf{0.62} \\
\hline
\multicolumn{3}{c}
{\textbf{YaleB}~\cite{ExtendedYaleDataset2001}} \\
\hline 
Method & Tar. Acc. ($\uparrow$) & Adv. Acc. ($\downarrow$)\\
\hline
ML-ARL~\cite{xie2017controllable} &  {0.89} & 0.57 \\
VFAE~\cite{louizos2017variationalfairautoencoder} & 0.85 & 0.57 \\
MaxEntARL~\cite{roy2019mitigating}  & 0.89 & 0.57 \\
Ker.-SARL~\cite{Sadeghi_2019_ICCV} & 0.86 & \textbf{0.20} \\
ODR~\cite{sarhan2020fairness}  & 0.84 & 0.52 \\
\hline
\textbf{Ours (\texttt{Focal entropy})}  & \textbf{0.90} & \textbf{0.20} \\
\hline
\end{tabular}
}
\caption{\textbf{Fair classification:}  Results apart from ours originate from \cite{roy2019mitigating,Sadeghi_2019_ICCV,sarhan2020fairness}}
\label{tab:results_fair}
\end{table} 

\begin{table*}[]
\centering
\resizebox{16.0cm}{!}{%
\begin{tabular}{l|ccc|ccc|ccc}
\hline
\multicolumn{1}{c|}{}                                                 & \multicolumn{3}{c|}{\textbf{CelebA~\cite{guo2016ms}}} & \multicolumn{3}{c|}{\textbf{Adult}~\cite{Dua:2019}} & \multicolumn{3}{c}{ \textbf{UTKFace}~\cite{zhifei2017cvpr}} \\ \hline
{Method}                                                       & {Tar. Acc. ($\uparrow$)}  & {EOD ($\downarrow$)}  & {PD ($\downarrow$)} & {Tar. Acc. ($\uparrow$)}   & {EOD ($\downarrow$)}   & {PD ($\downarrow$)}   & {Tar. Acc. ($\uparrow$)}      & {EOD ($\downarrow$)}     & {PD ($\downarrow$)}     \\ \hline
FADES~\cite{Jang_2024_CVPR}                   & 0.92                             & {0.03}                & 0.12                       & \textbf{0.85}                     & 0.10                          & 0.16                         & 0.80                                 & {0.06}                   & \textbf{0.10}                  \\
GVAE~\cite{ding2020guided}                      & 0.92                             & 0.05                         & 0.13                       & 0.85                              & 0.11                          & 0.18                         & {0.82}                        & 0.21                            & 0.20                           \\
FFVAE~\cite{creager2019flexibly}                & 0.89                             & 0.07                         & \textbf{0.07}              & 0.80                              & 0.06                          & \textbf{0.09}                & 0.77                                 & 0.27                            & 0.21                           \\
ODR~\cite{sarhan2020fairness}                   & 0.88                             & 0.04                         & 0.10                       & 0.79                              & 0.26                          & 0.16                         & 0.74                                 & 0.17                            & 0.21                           \\
FairDisCo~\cite{10.1145/3534678.3539302}        & 0.84                             & 0.07                         & 0.05                       & 0.80                              & 0.13                          & 0.14                         & 0.77                                 & 0.27                            & 0.20                           \\
FairFactorVAE~\cite{10.1007/s11063-022-10920-8} & 0.91                             & 0.05                         & 0.14                       & 0.78                              & 0.10                          & 0.13                         & 0.72                                 & 0.10                            & 0.14                           \\ \hline
\textbf{Ours (\texttt{Focal entropy})}               & \textbf{0.98}                    & \textbf{0.02}                & 0.14                       & \textbf{0.85}                     & \textbf{0.05}                 & 0.13                         & \textbf{0.85}                                 & \textbf{0.05}                            & {0.16}                  \\ \hline
\end{tabular}
}

\caption{\textbf{Fairness classification with fairness violation measures: } Results for various approaches w.r.t. fair classification. Results apart ours originate from~\cite{Jang_2024_CVPR}}
\label{tab:fairness}
\end{table*}

In order to fulfill the properties of an entropy, the $\pi_j$ have to be normalized subsequently. This is achieved according to $\pi_j\cdot(\sum_i^N \pi_i)^{-1}=\pi_j^*$.  Thus we yield the off-centered entropy $\eta(.)$ of probabilities $\mathbf{p}$ defined as $\eta(\mathbf{p})=h(\bm{\pi}^*)=-\sum_{j=1}^N \pi^*_j \log \pi^*_j$. Consequently, adversarial sanitization seeks to maximize the off-centered entropy w.r.t. private attribute on $\vz_\public$ according to:
\begin{equation}
 \phi_{\tilde{S}}(\mathbf{\theta}_E,\tilde\theta_\public) = \eta(\sigma_{\vz_\public}^{\tilde{S}}).
 \end{equation}

Here $\sigma_{\vz_\public}^{\tilde{S}} \in \mathbb{R}^N$ denotes the softmax vector of the adversarial predictor $\tilde{p}_S(s|\vz_\public; \tilde\theta_\public)$.
Maximizing the off-centered entropy is analogous to minimizing the Kullback-Leibler divergence w.r.t. $\bm{\tau}$, such that we yield:
 \begin{equation}
 \begingroup\color{OrangeRed}
     \phi_{\tilde{S}}(\mathbf{\theta}_E,\tilde\theta_\public) = D_{KL}(\tilde{p}_S({s}|\vz_\public)\|\bm{\tau};\tilde\theta_\public)
     \endgroup
 \end{equation}
 
\noindent\textbf{Optimization:}
 Instead of transforming the probabilities, we can emulate this by splitting the entropy computation into two parts. This is computationally faster and conceptually more straightforward. Then training involves maximization of entropy within each subgroup separately:
 \begin{equation}
 \begin{split}
      \phi_{\tilde{S}}(\mathbf{\theta}_E,\tilde\theta_\public) = D_{KL}(\tilde{p}_S(s_{s}|\vz_\public);\tilde\theta_\public)\|\mathcal{U}_{s}) + \\
      D_{KL}(\tilde{p}_S(s_{d}|\vz_\public);\tilde\theta_\public)\|\mathcal{U}_{d}) \label{eqn:split_entropy}
\end{split}
 \end{equation}
 
Here, $s_{s}, s_{d}$ denote the sensitive attributes w.r.t. the label subsets of the respective group, and $\mathcal{U}_*$ the associated uniform distribution. That is while maximizing the entropy w.r.t. similar group members, the same is done for the non-similar group members. This helps to support the information disentanglement. It should be noted that merging the groups ``\emph{similar}'' and ``\emph{dissimilar}'' results in Eq.~\ref{eqn:split_entropy} transform $D_{KL}(\tilde{p}_S(s|\vz_\public);\tilde\theta_\public)\|\mathcal{U})$, i.e., maximizing the uniform entropy across the entire sensitive attributes.

\section{Experiments \& Results}
\label{sec:experiments}
 \begin{table}
\centering
 \resizebox{8.5cm}{!}{%
\begin{tabular}{l|ll}
\hline
\multicolumn{3}{c}{\textbf{CIFAR-100}~\cite{Krizhevsky09learningmultiple}} \\
\hline
Method & Tar. Acc. ($\uparrow$) & Adv. Acc. ($\downarrow$)\\
\hline
$\uparrow$ Upper bound  & 1.0 & 0.01 \\
\hdashline
Ours (\texttt{entropy}) & 0.70 & 0.16 \\
\textbf{Ours (\texttt{Focal entropy})} & \textbf{0.82} & \textbf{0.16} \\
\hline
\multicolumn{3}{c}{\textbf{CelebA}~\cite{guo2016ms}} \\
\hline
Method & Tar. Acc. ($\uparrow$) & Adv. Acc. ($\downarrow$)\\
\hline
$\uparrow$ Upper bound / Chance & 1.0 & $<$ 0.001   \\
\hdashline
Ours (\texttt{entropy}) & 0.90 & 0.061 \\
\textbf{Ours (\texttt{Focal entropy})}  & \textbf{0.90} & \textbf{$<$ 0.01} \\
\hline
\end{tabular}
}
\caption{Ablation study on the choice of entropy.
}
\label{tab:ablation}
\end{table} 

\noindent\textbf{Implementation Details:} We follow a two-step training process, largely following~\cite{wu2020privacy}. This comprises an (i) “initial train” and (ii) “burn-in” phase, whereby each phase entails training for $\omega_{.}$ epochs. The warm-up stage prepares the encoder for generating meaningful representations (for utility),  i.e., disentangling the information. (ii) integrates the adversary, anonymizing the representation (for privacy). 
We employ ADAM optimizer with weight-decay of $\{1\mathrm{e}{-3}, 1\mathrm{e}{-4}\}$, with a learning rates of $\{0.01, 0.001\}$. The number of epochs was determined empirically based on the observation of loss convergence.
We adopted the following configurations for k-NN size: CIFAR:  $k=5$, CelebA: $k=16$. For the other datasets, we set $k$ equal to the number of sensitive attribute classes.
 Parameters for the trade-off Eq.~\ref{eq:cost_function} were selected with hyperparameter search, giving maximum priority to privacy sanitization over utility. Specifically, we obtained hyperparameters with a grid search $\alpha_*, \beta_* \in [0,1]$ to achieve the best target accuracy/adversarial accuracy ratio and PD, respectively.
For obtaining image reconstruction, we added a VAE reconstruction term to the loss. The training was conducted using NVIDIA A10 GPUs, with an aggregated training time of $\qty{4}{\hour}$ on CIFAR-100 and Adult, $\qty{1}{\day}$ on CelebA, and \textless $\qty{1}{\hour}$ for German and YaleB.

\subsection{Privacy Results}
\label{sec:privacy_results}
Given the impossibility of providing privacy guarantees against ``any'' unseen attacker model (e.g., oracle classifier; known as the $\forall$-\emph{challenge} - discussed in~\cite{wu2020privacy}), and for the sake of comparability, we follow the evaluation protocol of~\cite{roy2019mitigating,sadeghi2019global}. This entails training a ``probing'' classifier that acts as a proxy for the oracle classifier.  
We evaluated the privacy sanitization in terms of target and adversarial accuracy on two benchmarks. Additionally, we performed a probing experiment to assess the classifier capacity's dependency regarding the learned representation's privacy leakage potential. Specifically, we determined the adversarial accuracy on the learned representation wrt. to the classifier's strength. The \texttt{strong} probing classifier has approximately double the proxy classifier's capacity used for training -  Tab.~\ref{tab:prob}. Although agnostic in terms of architecture, we selected a VAE~\cite{kingma2013autoencoding} with 512 latent dimensions for $\vz_\public$ and $\vz_\public$, respectively. Employing VAE provides us with a rich representation learning framework in conjunction with interpretability.

\noindent\textbf{Privacy Sanitization on CIFAR-100: } As the first testbed, we adopt the ``simulated'' privacy problem proposed by \cite{roy2019mitigating} designed on the CIFAR-100 dataset. The dataset consists of 100 classes grouped into 20 superclasses. 
We treat the \emph{coarse} (superclass) and \emph{fine} (class) labels as the target and sensitive attribute, respectively. The task is to learn the superclasses' features while not revealing the information about the underlying classes. Here we note that ``ideally'' we desire a predictor accuracy of 100\%, an adversary accuracy of 1\% (random chance for 100 classes).  In Tab.~\ref{tab:results_privacy}, we report the accuracy achieved by the attribute predictor and adversary. 
From these results, we observe that with our proposed method, the representation achieves the best target accuracy while being comparable to adversary accuracy compared to the SOTA sanitization methods \cite{roy2019mitigating,Sadeghi_2019_ICCV}. Using conventional entropy for adversary instead of our proposed \emph{focal entropy} results in a significant performance drop (from 82\% to 70\% in terms of target accuracy at comparable adv. accuracy). This indicates the importance of the integration of class similarity in the privacy sanitization step.

In Fig.~\ref{fig:tradeoff_target}, we report the characteristics of the proposed approach in terms of trade-off curves that portray the correlation between privacy and target utility. We compared the most relevant works to us: MaxEnt-ARL and ML-ARL from \cite{roy2019mitigating}, Kernel-SARL \cite{Sadeghi_2019_ICCV}, Orthogonal Disentangled Representations (ODR) from \cite{sarhan2020fairness}, and vanilla baseline \emph{(No Privacy)} without sanitization. As can be seen, the higher the correlation to privacy, the higher the loss of accuracy at a high level of sensitive accuracy appears. Furthermore, the proposed approach features a significantly better utility trade-off in the high target accuracy domain.  It should be noted that the emphasis of the proposed approach is on the accuracy gain achieved in the high utility domain (highlighted with the dashed gray line in Fig.~\ref{fig:tradeoff_target}). 
Additionally, to assess the effect of our proposed \emph{focal entropy} in terms of the adversarial component, comparing it with conventional entropy. We report the results in Tab. \ref{tab:ablation}.

\noindent\textbf{Privacy Sanitization on CelebA: } 
As the second testbed, we adopt the CelebA dataset, which has a richer structure to be utilized for privacy while being less structured in terms of similarity. 
It consists of more than 200k images from $10.177$ identities along with 40 binary attributes. Specifically, we treat the attribute labels as target and celebrity identity (i.e., ID) as the source of sensitive information. Simultaneously, identities are by design in a joint space. The task is to learn to classify the attributes while not revealing the information about the identities. We note that ideally, we desire a lower adversary accuracy ($<$ 0.001) compared to the previous case, as the number of ID classes is an order of magnitude higher. In Tab.~\ref{tab:results_privacy} we report the target/adversary accuracy. We observe that the representation we learn achieves a higher target accuracy and lower adversary accuracy than the strong baseline ML-VAE~\cite{bouchacourt2018multi}. Using standard entropy instead of our proposed \emph{focal entropy} in CelebA shows a considerably higher privacy leakage (namely, 0.061 and $<$ 0.01) - see~Tab.~\ref{tab:ablation}.

\begin{figure}[b!]
\centering
\includegraphics[width=0.45\textwidth]{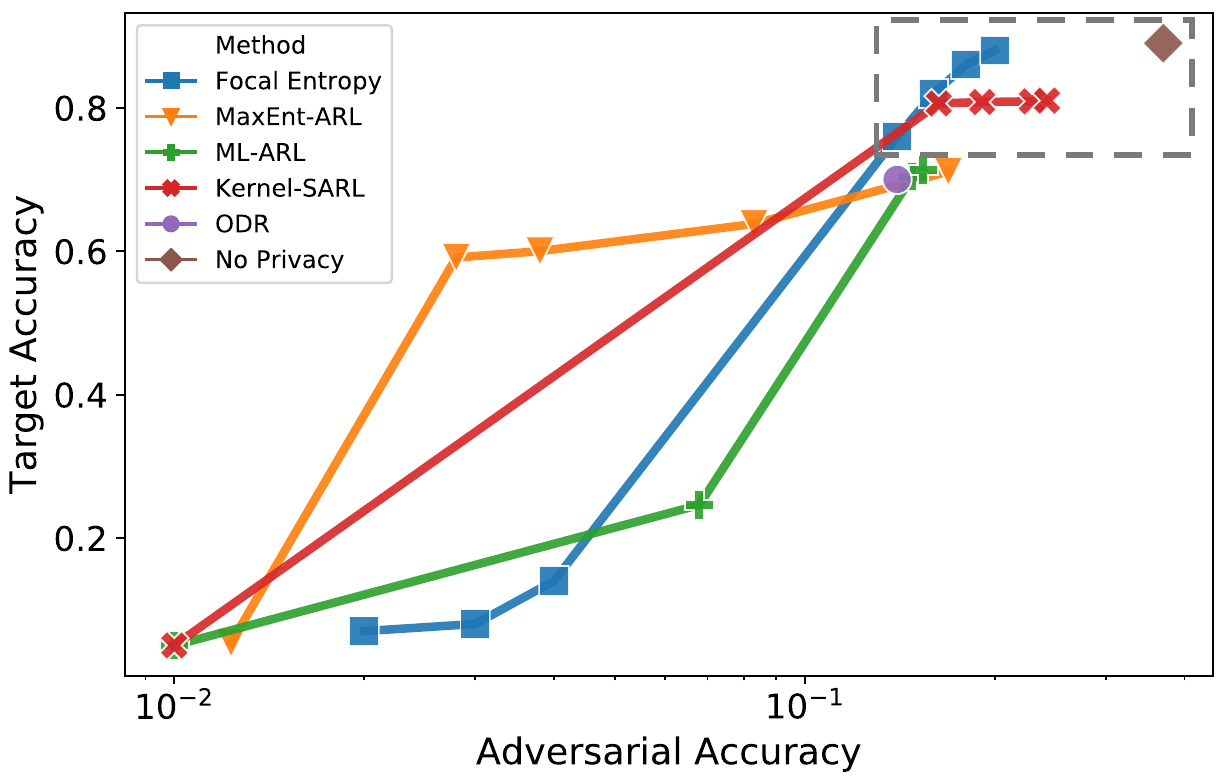}
\caption{Trade-off curve between target and adversarial accuracy on CIFAR-100 (dashed box denotes the ROI).
}
\label{fig:tradeoff_target}
\end{figure}

\noindent\textbf{Privacy-Utility Trade-Off: }\label{sec:tradeoff_curve} Characteristics of the proposed approach in terms of complete trade-off curves that portray the correlation between privacy and target utility. We compared the most relevant works to us: MaxEnt-ARL and ML-ARL from \cite{roy2019mitigating}, Kernel-SARL \cite{Sadeghi_2019_ICCV}, Orthogonal Disentangled Representations (ODR) from \cite{sarhan2020fairness}, and vanilla baseline \emph{(No Privacy)} without sanitization. The gray dashed box denotes the region of interest in terms of maximum utility.

\begin{table}
\centering
 \resizebox{8.5cm}{!}{%
\begin{tabular}{l|ll}
\hline
\multicolumn{3}{c}{\textbf{CelebA}~\cite{guo2016ms}} \\
\hline
Method & Tar. Acc. ($\uparrow$) & Adv. Acc. ($\downarrow$)\\
\hline
$\uparrow$ Upper bound / Chance & 1.0 & $<$ 0.001   \\
\hdashline
Ours (\texttt{normal cls.}) & 0.90 & 0.007\\
Ours (\texttt{strong cls.}) & 0.90 & 0.009 \\
\hline
\end{tabular}
}
\caption{Probing analysis on CelebA.
}
\label{tab:prob}
\end{table}

\begin{figure*}[t!]
\centering
\includegraphics[width=0.72\textwidth]{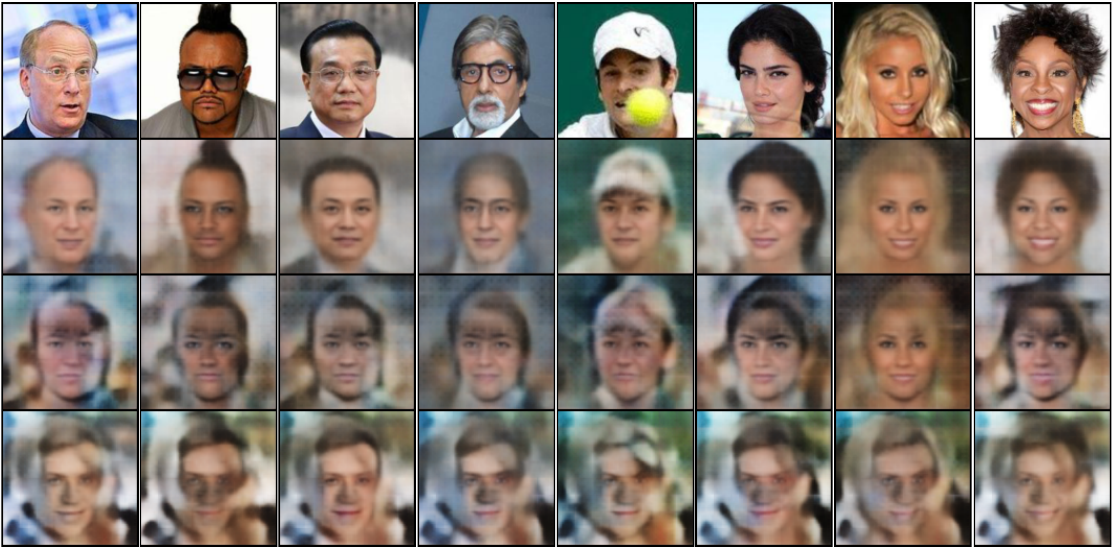}
\caption{Visualization of CelebA data and reconstructions at different privacy levels. From top to bottom, privacy revelation is decreasing.}
\label{fig:facerecon}
\end{figure*}

\subsection{Analysis and Discussion}
\noindent\textbf{ Attribute-level Privacy Analysis: }\label{sec:attribute-privacy}

\begin{figure}
\centering
\includegraphics[width=0.5\textwidth]{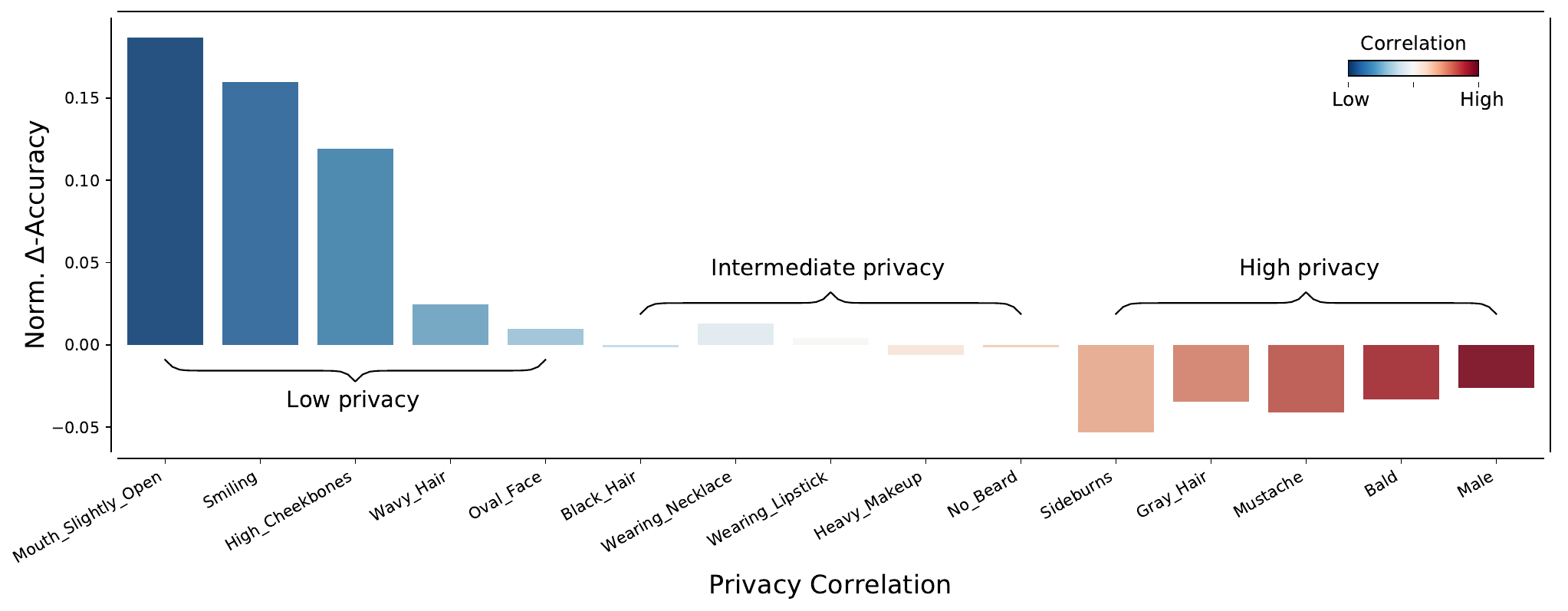}
\caption{\textbf{Attribute-level privacy analysis:} The normalized $\Delta$-Accuracy and privacy trade-off on CelebA.}
 \label{fig:privacyanalsys}
\end{figure}

We analyze the attribute classification accuracy behavior w.r.t. varying levels of privacy of attributes. To our knowledge, this is the first exploration of attribute-level privacy-preserving representation learning algorithms on the Celeb-A dataset. The rich correlation structure amongst all attributes makes this a challenging privacy dataset; i.e., it is difficult to achieve high accuracy for non-private attributes and low accuracy for private ones. Therefore representative subsets of the original attributes were chosen for more in-depth analysis (see appendix for complete results on attribute-level privacy). The subsets were chosen on the grounds of different values of correlation with IDs, reflecting the different privacy levels of each. Each of the three subsets (low, intermediate, and high privacy, resp.) consists of 5 attributes. 
Data statistics reported comprise normalized accuracy (w.r.t. prior) of the classifiers trained on the target representation $\vz_\public$, the $\Delta$-accuracy for each attribute corresponding to the difference of the accuracy between two classifiers — the one trained on $\vz_\public$, and the one trained on $\vz_\private$. For a visualization of the privacy-attribute correlation, see Fig.~\ref{fig:privacyanalsys}. As can be seen, the higher the correlation to privacy, the higher the loss in accuracy appears. Privacy-revealing attributes such as gender and facial attributes strongly correlate, therefore exhibiting a significant loss in accuracy. In contrast, attributes related to temporal facial features, such as gestures, feature low privacy correlation and, therefore, achieve strong accuracy. Conversely, the more unrelated a variable is to the identity, the higher gains in accuracy manifest.  

\noindent\textbf{Ablation Analysis on Loss Components: }\label{sec:loss_ablation}
To assess the contribution of each component of our objective function, we evaluated each module's performance separately, gradually adding components. Ablation was conducted with VAE and an additional reconstruction loss component: \emph{reconstruction loss, target classification loss, adversary loss}. For that, we report the ablation study of the loss components on CelebA in Tab. \ref{tab:ablation_loss}.

\noindent\textbf{t-SNE Visualization: }\label{sec:tsne}
 To analyze the privacy sanitization in the latent representation, we visualize the t-SNE of the target and residual parts in Fig.~\ref{fig:tsne}. For visualization, a subset of 200 IDs was chosen randomly from the test set. As can be seen, the private class associations appear random and cannot be recovered from the target part. In contrast, the residual part allows the formation of clusters for common IDs. 

\subsection{Qualitative Results}\label{sec:qual}
We show some qualitative results of our method in Fig.~\ref{fig:facerecon}. 
From top to bottom, privacy disclosure is decreasing. Each column is a different sample from CelebA. The first row corresponds to the ground truth image. The second row corresponds to reconstruction from latent representation consistent with the concatenation of target and residual content. The last two rows correspond to the reconstruction from the target and residual of the learned representation. Specifically, the third row shows the reconstruction from the concatenation of the average target latent (across the entire training set) and the residual latent representation. The fourth row corresponds to the reconstruction from the target latent representation, assuming a zero private latent vector. It can be noticed that visualizations from target latent and residual parts confirm the sanitization visually. More concretely, reconstructions from the residual part of the representation result in a sharpening of identity, revealing properties such as skin color and eye style. Simultaneously, in these visualizations, one observes the vanishing public attributes such as sunglasses and/or jewelry.
In contrast, reconstructions from the target representation contain generic facial features, e.g., smiles, teeth, and face outlines. Specifically, it should be noted that target reconstructions usurp properties provided by the CelebA attributes, e.g., wearing glasses (shadows around eyes), smiling, etc. Residual reconstructions feature essential information to identify the person, e.g., beard, eyeglasses, or background removed.

\subsection{Fair Classification}\label{sec:fairness}
We evaluate fair downstream classification on diverse datasets, including CelebA~\cite{guo2016ms} (details see above in Sec.~\ref{sec:privacy_results}) and UTKFace~\cite{zhifei2017cvpr} (over 20,000 faces between in a range of 0 to 116 years old people) for facial attribute classification and the Adult~\cite{Dua:2019} income dataset for tabular data. Following previous works~\cite{Jang_2024_CVPR,9878694,10.1007/978-3-030-65414-6_35,10021090}, we use CelebA to predict the ``Smiling'' (c.f. ``identity'' in the privacy setup above) attribute and UTKFace to classify individuals as over 35 years old, both using gender as the sensitive attribute. The Adult income dataset aims to predict income exceeding \$50K, again with gender as the sensitive attribute. Finally, the German credit dataset comprises 1000 instances, each with 20 attributes. The dataset aims to classify bank account holders as having good or bad credit, with gender identified as a sensitive attribute.
For a comprehensive and fair comparison, we conducted experiments following the evaluation setup and protocols of ~\cite{roy2019mitigating,sadeghi2019global} and ~\cite{Jang_2024_CVPR}, respectively. The former employs different MLP-encoder networks and partially leverages pre-computed ResNet-100 features for vision tasks, reporting target, and adversarial accuracy. The latter employs a ResNet-18~\cite{he2016residual} encoder architecture for vision and an MLP for tabular tasks. Besides target accuracy, fairness violations are measured using established metrics, including demographic parity (DP)~\cite{barocas2014datas} and equalized odds (EOD)~\cite{10.5555/3157382.3157469}. Experimental results demonstrate the efficacy of focal entropy in balancing classification accuracy and privacy/fairness. Focal entropy consistently attained leading accuracy with privacy and fairness violation levels comparable to state-of-the-art methods, often outperforming them. In terms of experiments w.r.t. setup of ~\cite{roy2019mitigating,sadeghi2019global} (see Tab.~\ref{tab:results_fair}), focal entropy shows consistent best performance often with a significant margin over approaches such as MaxEnt-ARL~\cite{roy2019mitigating}, Kernel-SARL and~\cite{Sadeghi_2019_ICCV}, ODR~\cite{sarhan2020fairness}. In terms of experiments w.r.t. the setup of~\cite{Jang_2024_CVPR} (see Tab.~\ref{tab:fairness}), FFVAE~\cite{creager2019flexibly}, ODR~\cite{sarhan2020fairness}, FairDisCo~\cite{10.1145/3534678.3539302} and FairFactorVAE~\cite{10.1007/s11063-022-10920-8} exhibited mixed performance. FADES~\cite{Jang_2024_CVPR} and GVAE~\cite{ding2020guided} achieved results comparable to focal entropy. Notably, our proposed method consistently surpassed all methods across all datasets at the accuracy level and EOD fairness measure. This success validates the effective disentanglement of correlated features, yielding superior latent representations through successful feature isolation. This implies the preservation of information pertinent to the target label while mitigating leakage.

\begin{table}
\centering
 \resizebox{8.5cm}{!}{%
\begin{tabular}{l|ll}
\hline
\multicolumn{3}{c}{\textbf{CelebA}~\cite{guo2016ms}} \\
\hline
Method & Tar. Acc. ($\uparrow$) & Adv. Acc. ($\downarrow$)\\
\hline
$\uparrow$ Upper bound  & 1.0 & $<$ 0.001   \\
\hdashline
Ours (only \emph{Rec.} loss) & 0.88 & - \\
Ours (\emph{Rec. + Tar.} loss) & 0.91 & 0.751 \\
Ours (\emph{Rec. + Tar. + Adv.} loss) & 0.90 & $<$ 0.01 \\
\hline
\end{tabular}
}
\caption{Ablation analysis for loss components on CelebA. }
\label{tab:ablation_loss}
\end{table}
\section{Conclusion}
\label{sec:conclusion}
This paper proposes an adversarial representation learning method to deal with a large overlap between the target and sensitive attributes. Training the representation entails decomposition into target and residual parts. Here, the target part is shareable without privacy infringement and facilitates applicability for a target task. In contrast, the residual part subsumes all the private non-shareable information. Our proposed method employs focal entropy for the privacy sanitization of the representation. The proposed approach is conceptually simple and architecture-agnostic. As the experiments suggest, the proposed approach is largely equal to state-of-the-art approaches, occasionally superior to the competitors, confirming that our method learns sanitized representations using less supervision. The current work considers only 1-hot vector sensitive attributes, as it is a more natural scenario for the privacy problem. Extension to the vector of sensitive attributes is possible. 
As future work, we envisage exploring the zero-shot setup (unseen IDs), which is naturally permitted by the class-agnostic nature of entropy we adopted.

{
    \small
    \bibliographystyle{ieeenat_fullname}
    \bibliography{main}
}

\clearpage
\setcounter{page}{1}
\maketitlesupplementary

\begin{figure}[ht!]
\centering
\includegraphics[width=0.4\textwidth]{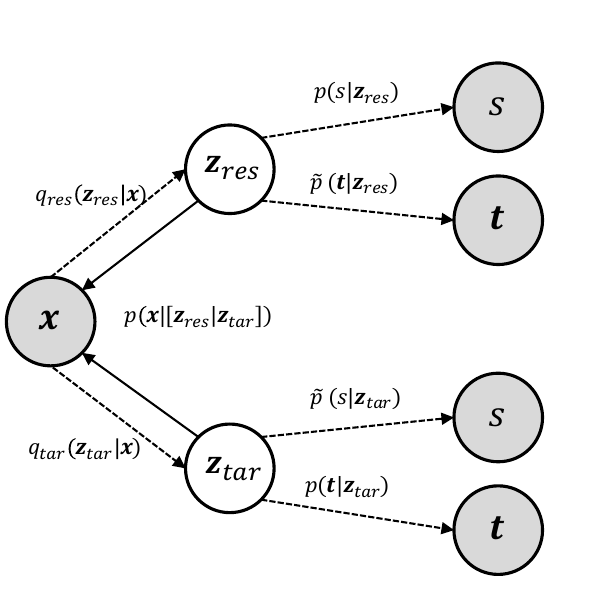}
\caption{\textbf{Schematic illustration of the proposed approach:} The graphical model associated with the minimax game with the latent variable split up in two components.}
\label{fig:method}
\end{figure}

In the following sections, we add additional details omitted in the main paper due to space restrictions. For clarity, we added a schematic illustration of the proposed approach - see Fig.~\ref{fig:method}.
In the first section, the sanitization convergence behavior of different classifiers involved in the adversarial minimax game is analyzed.
In the second section, we analyze the impact of varying the neighborhood size of $k-$NN in focal entropy on the adversarial accuracy with experiments conducted on the CelebA dataset.
In the third section, we analyze the effect of the \emph{classifier-strength} on the privacy leakage and dependence on training time.
We, then, present more visualizations around the concept of \emph{hub formation} and the connection to focal entropy. It contains a visualization of hub-forming identities and a zoom-in visualization of the adversarial remapping of the identities in CelebA. 
We also present detailed results of the accuracy and privacy trade-off on \emph{all} the attributes in CelebA dataset.
Additionally, we show a zoom-in version of the region of interest in the trade-off curve.
Next, we generated reconstructions on CelebA dataset samples.
Finally, \emph{architectural details} are presented in the last section.

\begin{figure*}[b!]
\begin{minipage}{.32\textwidth}
  \includegraphics[width=.99\linewidth]{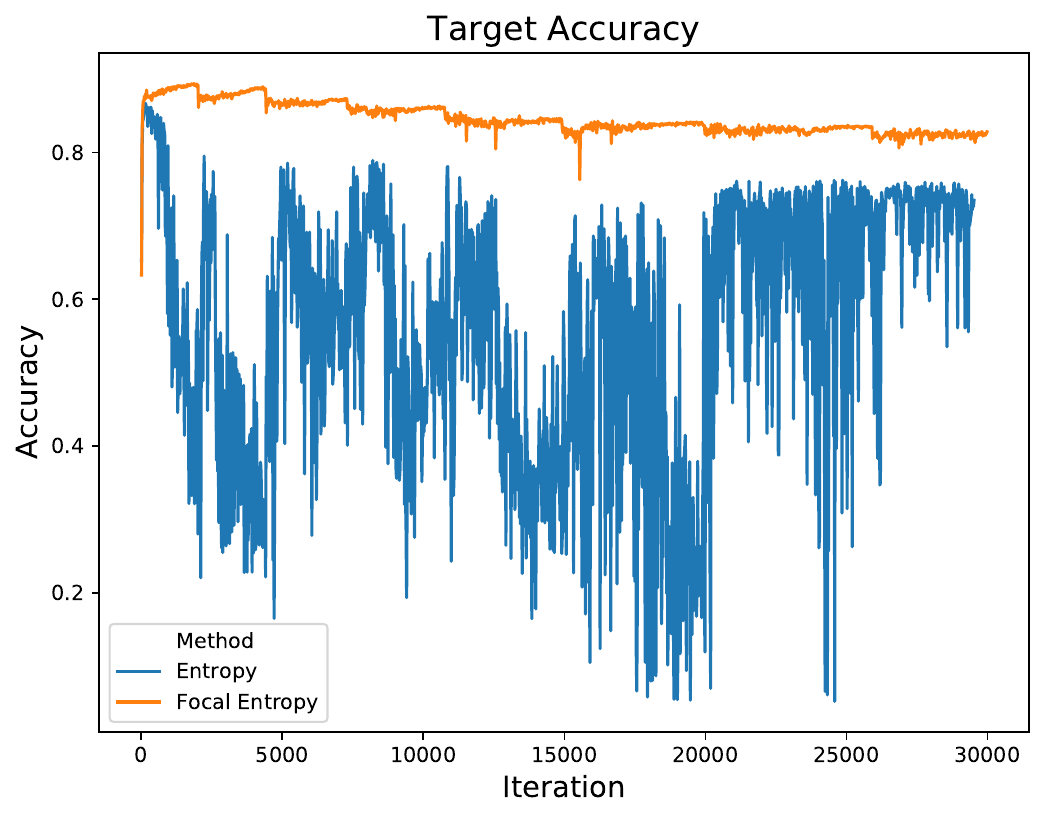}
\end{minipage}%
\begin{minipage}{.32\textwidth}
  \includegraphics[width=.99\linewidth]{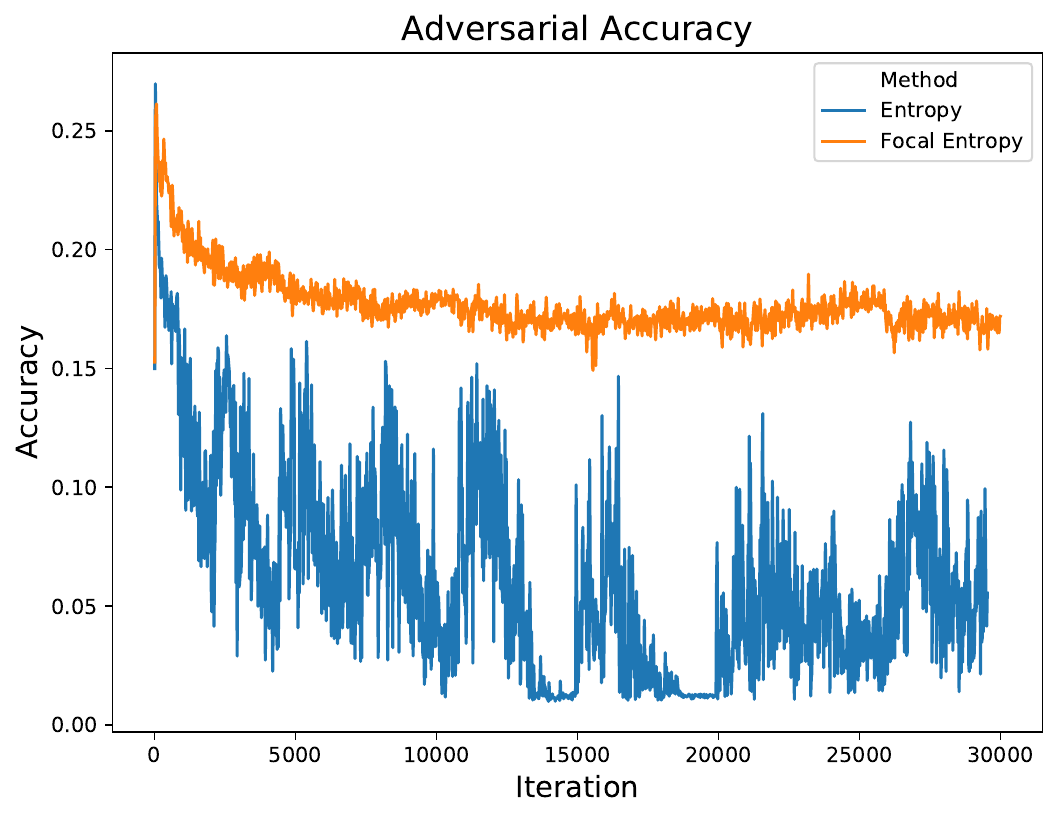}
\end{minipage}
\begin{minipage}{.32\textwidth}
  \includegraphics[width=.99\linewidth]{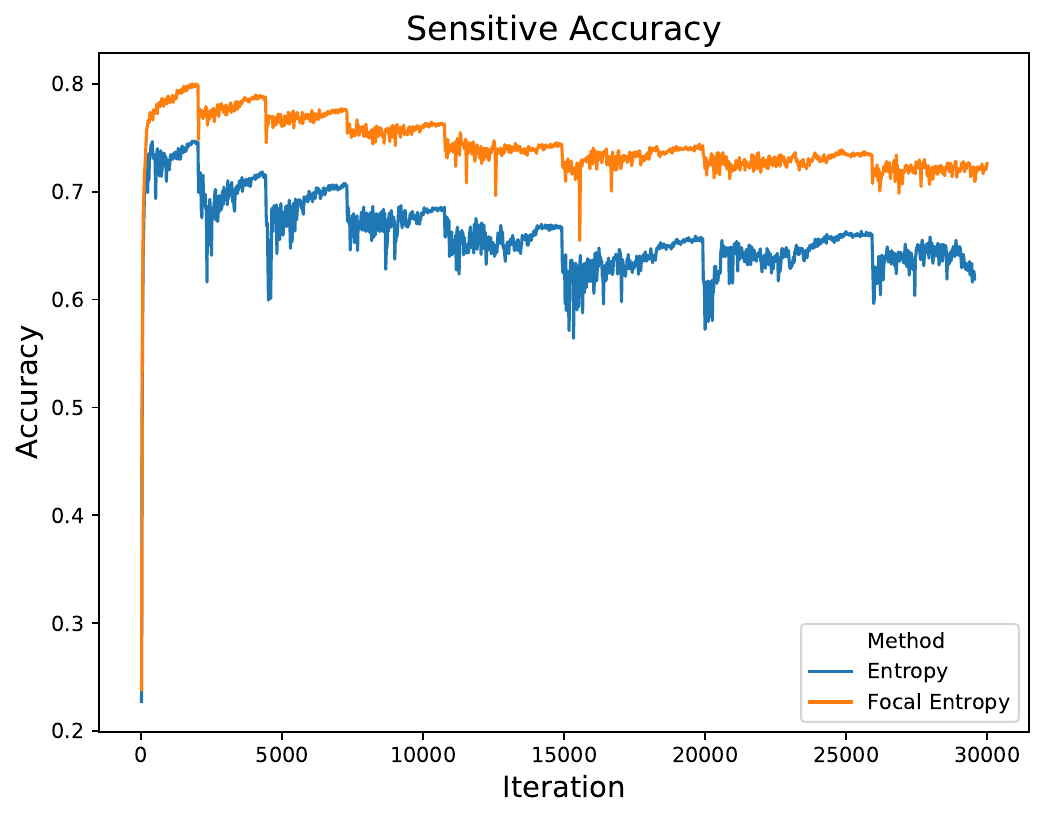}
\end{minipage}
\caption{Sanitization convergence behavior of standard entropy and focal entropy on CIFAR-100 for different classifiers: \textbf{Left:} Target accuracy, \textbf{Center:} Adversarial accuracy, \textbf{Right:} Sensitive attribute accuracy}
\label{fig:convergence}
\end{figure*}

\section{Sanitization Convergence Behaviour}\label{sec:convergence}

This section explores the behavior of standard entropy and the proposed focal entropy for sanitization. Fig.~\ref{fig:convergence} depicts the classification performance during the training of different classifiers involved in the minimax optimization scheme: target classifier accuracy, adversarial sensitive attribute accuracy, and sensitive attribute accuracy. As can be seen, employing standard entropy for sanitization results in re-occurring patterns of oscillations. This can be attributed to degenerate/trivial solutions and  ``shortcuts''. In contrast to that, focal entropy shows a relatively smooth convergence behavior. More details is in the main paper.

\section{Analysis on Neighborhood Size: }\label{sec:neighborhood_size}
We study the effect of varying $k$ on focal entropy and the associated adversary accuracy. See Fig.~\ref{fig:nn_r} for a visualization of this relationship on the CelebA dataset. As can be seen, the adversary accuracy has oscillatory behavior with various local minima, reaching optimum around $k=16$. This is attributed to the formation of information ``hubs''.

\section{Probing Analysis with Strong Classifier}\label{sec:cls_ablation}

\begin{figure*}
\hfill
\subfloat[Adversarial classifier training time dependency]{\includegraphics[width=0.45\textwidth]{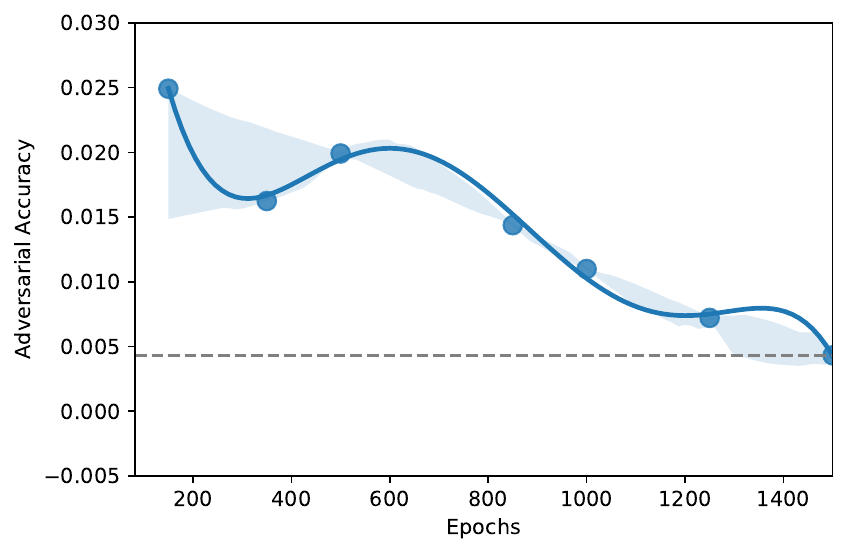}
\label{fig:iter-dependency}}
\hfill
\subfloat[Normal and strong adversarial classifier training time dependency]{\includegraphics[width=0.45\textwidth]{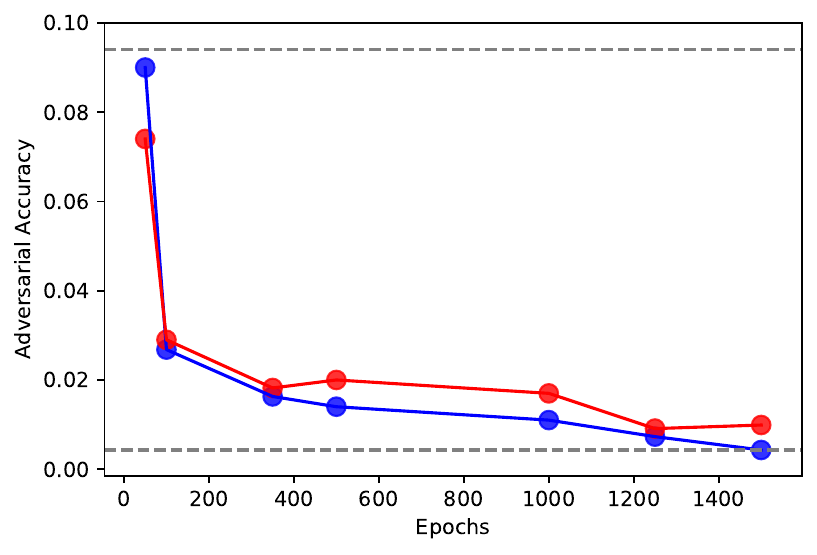}
\label{fig:iter-dependency-st}}
\hfill
\caption{\textbf{Left:} Relationship between adversarial accuracy and the number of training epochs on CelebA. The translucent band corresponds to 50\% confidence  minimum and maximum adversarial accuracy, respectively. \textbf{Right:} Relationship between adversarial accuracy for strong (red) and normal classifier (blue) w.r.t. the number of training epochs on CelebA. The translucent band corresponds to a 50\% confidence interval. Dashed lines correspond to the minimum and maximum adversarial accuracy, respectively.}
\label{fig:iter-dependency-strong}
\end{figure*}

\begin{figure}[ht!]
\centering
\includegraphics[width=0.45\textwidth]{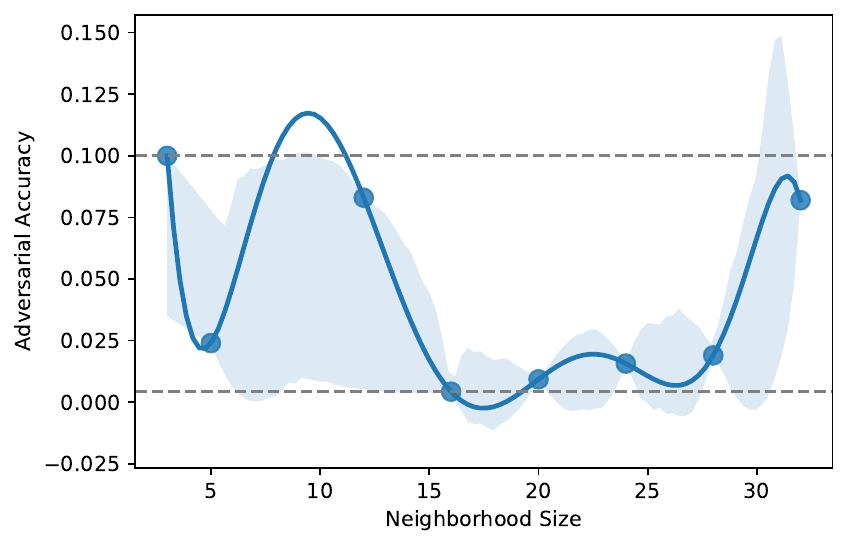}
\caption{ Neighborhood Analysis. Relationship between adversary accuracy and $k-$NN size on CelebA dataset. The translucent band corresponds to a 50\% confidence interval. Dashed lines correspond to a minimum, and maximum adversary accuracy, respectively.}
\label{fig:nn_r}
\end{figure}

This section provides more detail on assessing the classifier's strength in terms of privacy leakage and the dependence on training time. We thereby largely follow the protocol of~\cite{Jha_2018_ECCV,sadeghi2019global}. Specifically, we employed a \emph{stronger} post-classifier (Tab. 3 in the main paper) compared to the one used for learning the representation. The stronger post-classifier is endowed with additional layer stack (see Tab.~\ref{tab:strong_id_classifiers} for architectural details), trained for $100$ epochs.
The results in Tab. 3 of the main paper suggest no significant changes in target and adversarial accuracy. Figure~\ref{fig:iter-dependency-strong} extends these results, depicting the relationship w.r.t. the number of epochs. As can be seen, the difference between the \emph{strong} and \emph{normal} classifier is largely constant, independent of the epoch.

\section{Hub Analysis}\label{sec:hubanalysis}
\label{sec:vishub}

\begin{figure}[ht!]
\centering
\includegraphics[width=0.5\textwidth]{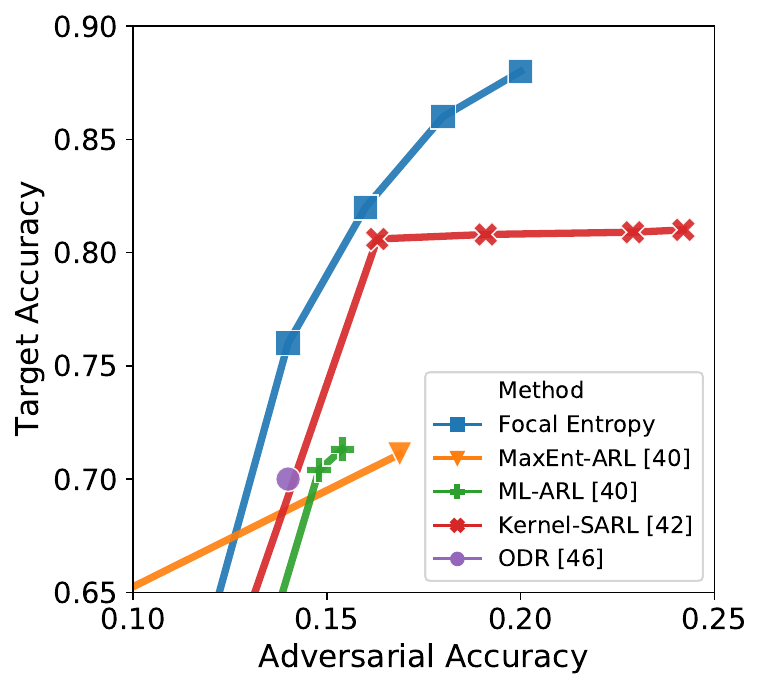}
\caption{Trade-off curve between target accuracy and adversarial accuracy on CIFAR-100 in the region of interest.}
\label{fig:tradeoff_target}
\end{figure}

\begin{figure*}[ht!]
\centering
\begin{minipage}{.24\textwidth}
  \centering
  \includegraphics[width=.99\linewidth]{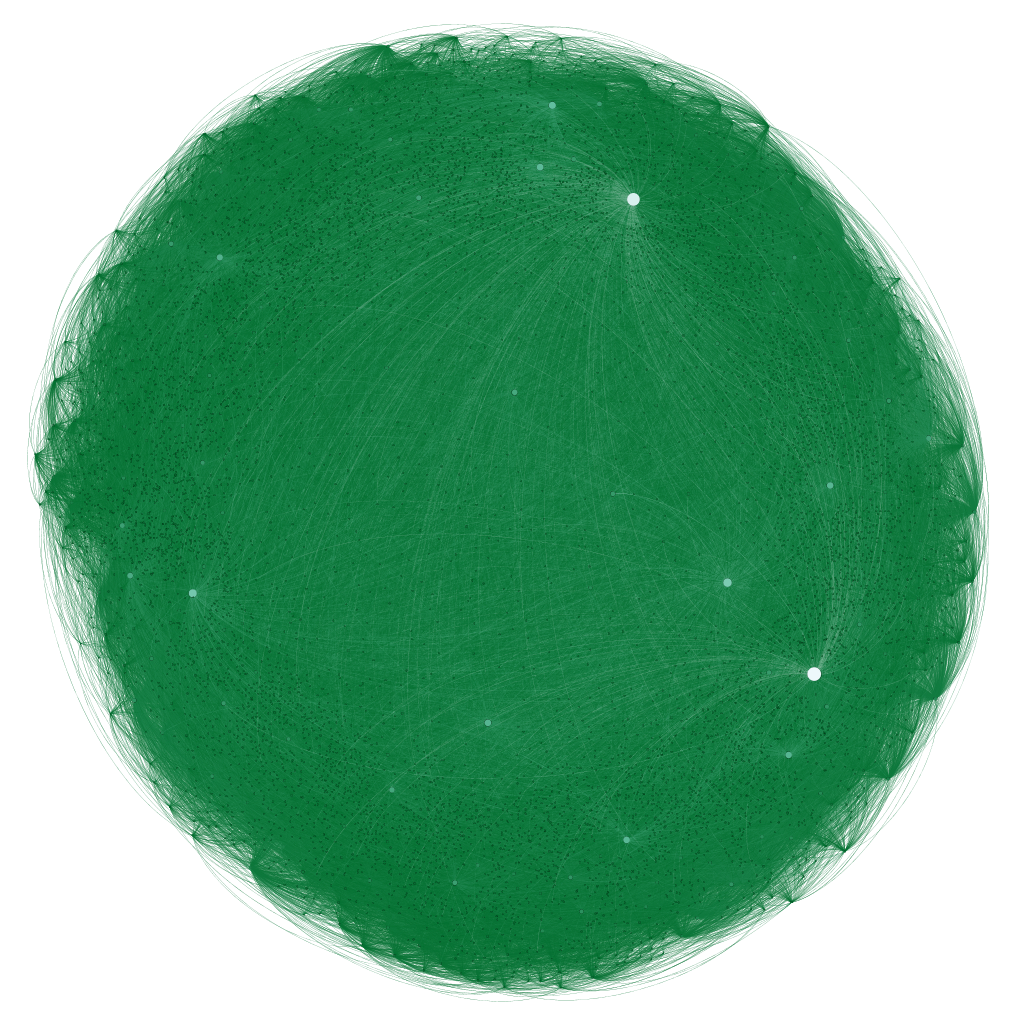}
\end{minipage}%
\begin{minipage}{.24\textwidth}
  \includegraphics[width=.99\linewidth]{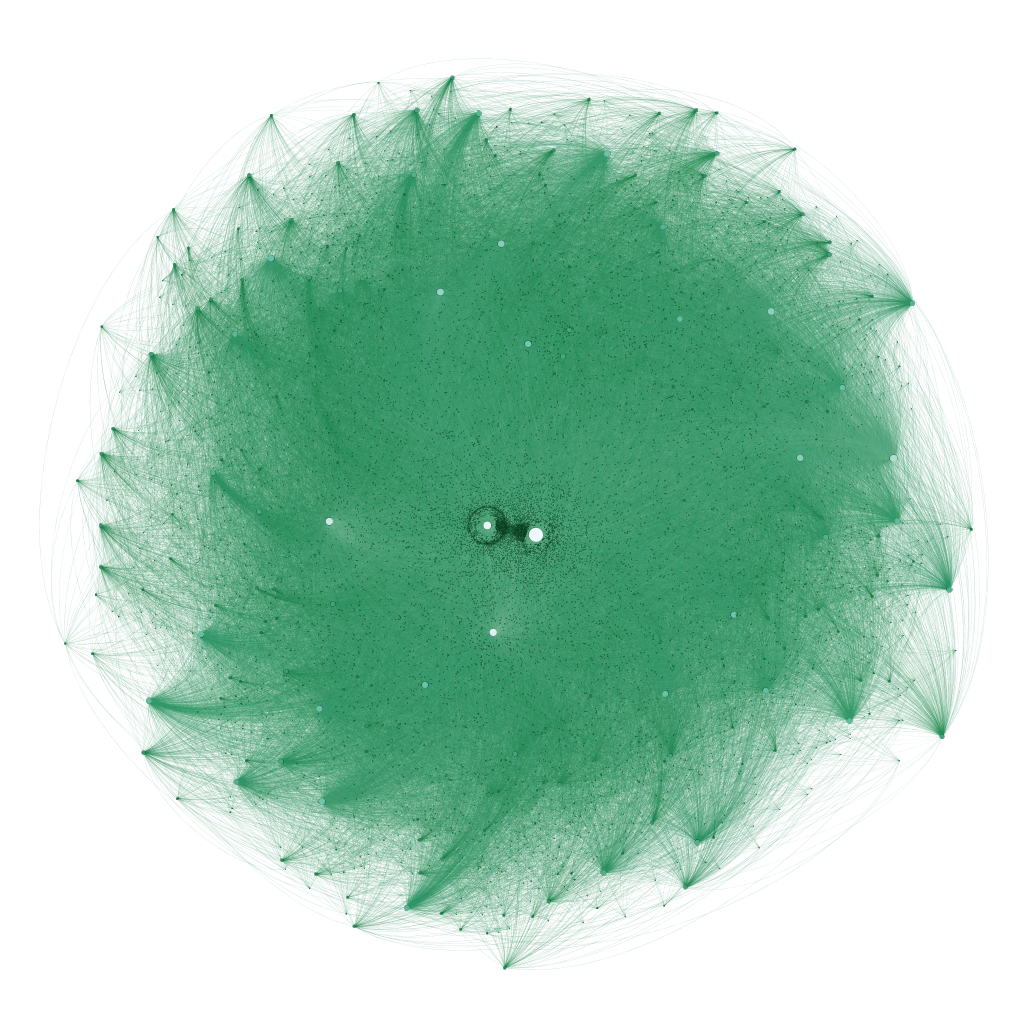}
\end{minipage}
\begin{minipage}{.24\textwidth}
  \includegraphics[width=.99\linewidth]{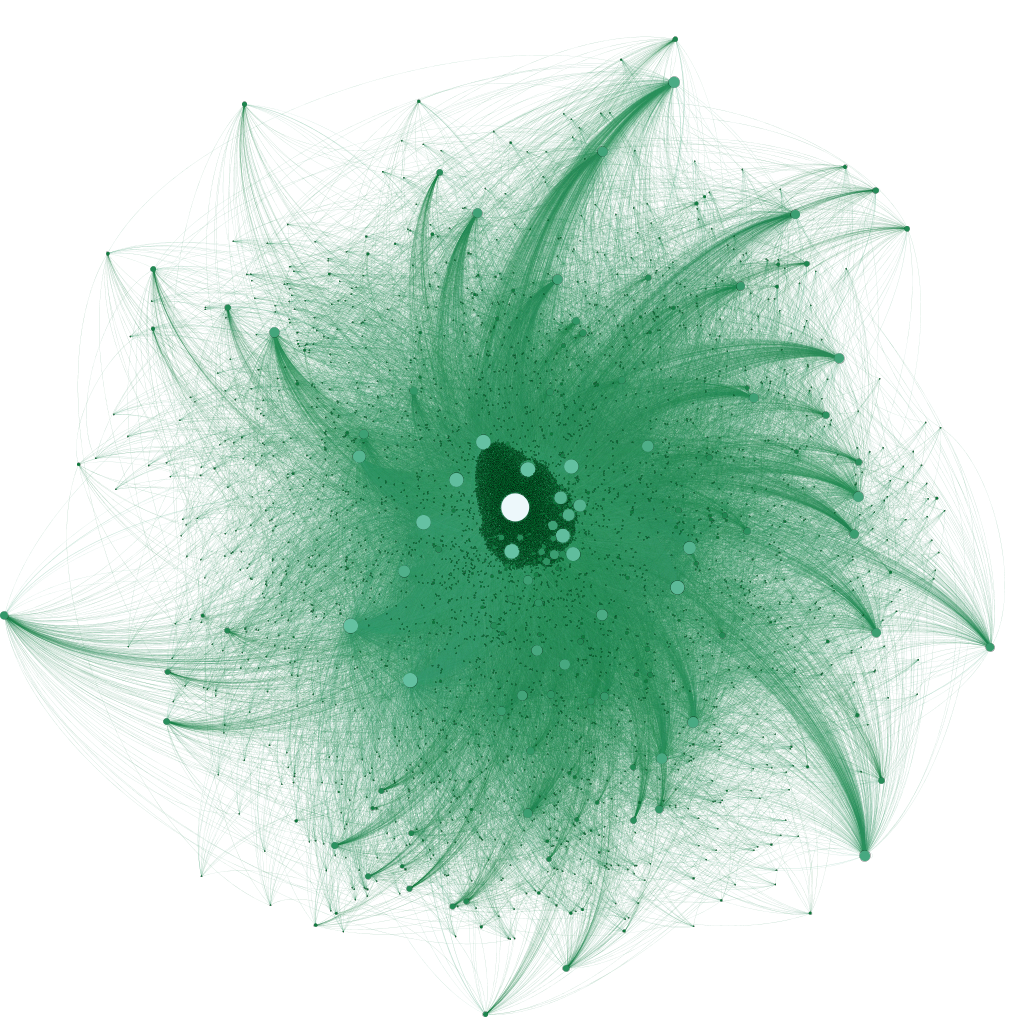}
\end{minipage}
\begin{minipage}{.24\textwidth}
  \includegraphics[width=.99\linewidth]{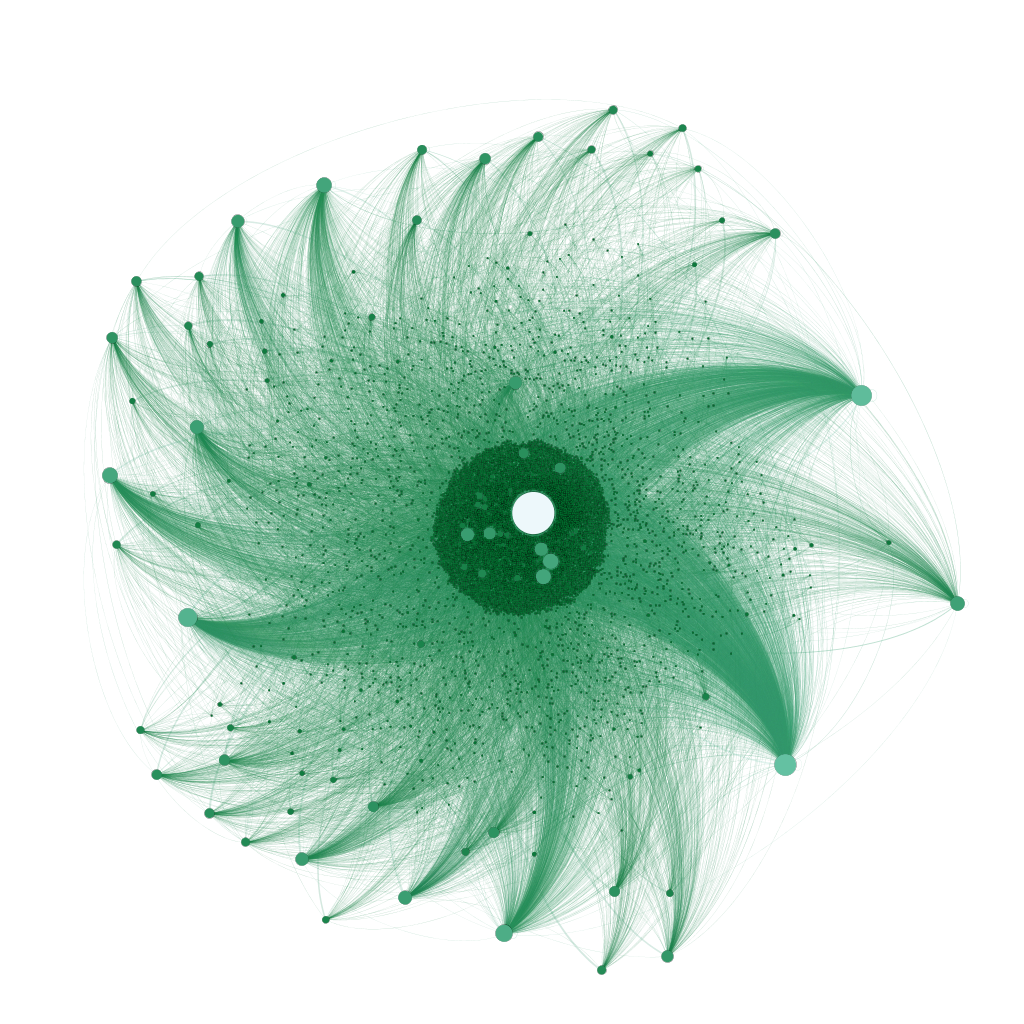}
\end{minipage}
\caption{Visualization of adversary ID re-mapping graph on CelebA for entropy to focal entropy with different k-NNs on $\vz_{\public}$. Nodes correspond to IDs, edges to adversarial re-mapping of an ID to facilitate adversary confusion. Node size/brightness scales with the number of associations (the bigger/brighter, the more IDs are mapped to a specific node). From \emph{left} to \emph{right}, increasing $k$ for focal entropy: $k=$1 ($\approx$ standard entropy), 2, 16 and 64.}
\label{fig:confusionnetwork}
\end{figure*}

This section provides an analysis of how the application of focal entropy, with its integration of the notion of $k-$NN, promotes the formation of ``hubs''. 
By varying the neighborhood size $k$, focal entropy manifests itself between two extremes: \textbf{i)} Choosing a \emph{small} $k$: Focal entropy in the limit approaches entropy  - no hubs are formed - remapping is a ``\emph{one-to-one}'' correspondence. This can be attributed to conventional entropy being invariant to random relabeling, 
giving rise to information preservation. This is prone to sub-optimal solutions since the forced equalization of classes by disregarding semantic similarity of classes makes the model susceptible to finding shortcuts such as label swapping. In this regard, target accuracy oscillations co-occurring with degenerate ID-remappings such as uniformity across all labels or temporary collapse to very few hubs were observed. \textbf{ii)} Choosing a \emph{large} $k$: Focal entropy promotes the formation of a single dominant hub, which is also referred to as the classical hubness problem. This phenomenon is related to the convergence of pairwise similarities between elements to a constant as the space's dimensionality increases~\cite{JMLR:v11:radovanovic10a} (collapsing on a single hub/trivial solution). The single hub then manifests itself as the favored result of queries~\cite{dinu2014improving} - giving rise to trivial solutions rather than deep sanitization. Choosing a \emph{non-extreme} $k$ leads to the formation of numerous equisized hubs, inducing a surjective mapping. As multi-hubs' formation coincides with a collapse, less information is required to establish a mapping. Hence, this characteristic gives rise to information removal and is thus imperative for proper sanitization.
Figure~\ref{fig:confusionnetwork} depicts the graphs induced by remapping for three different neighborhood configurations. Analyzing the nodes' average degree in the corresponding graphs, we observe a continuous decrease with growing neighborhood size $k$. Specifically, starting with entropy and increasing the increasing $k$ for focal entropy, we yield the average degrees of: $13.91$, $7.2$, and $3.0$.

\begin{figure*}[ht!]
\centering
\includegraphics[width=0.95\textwidth]{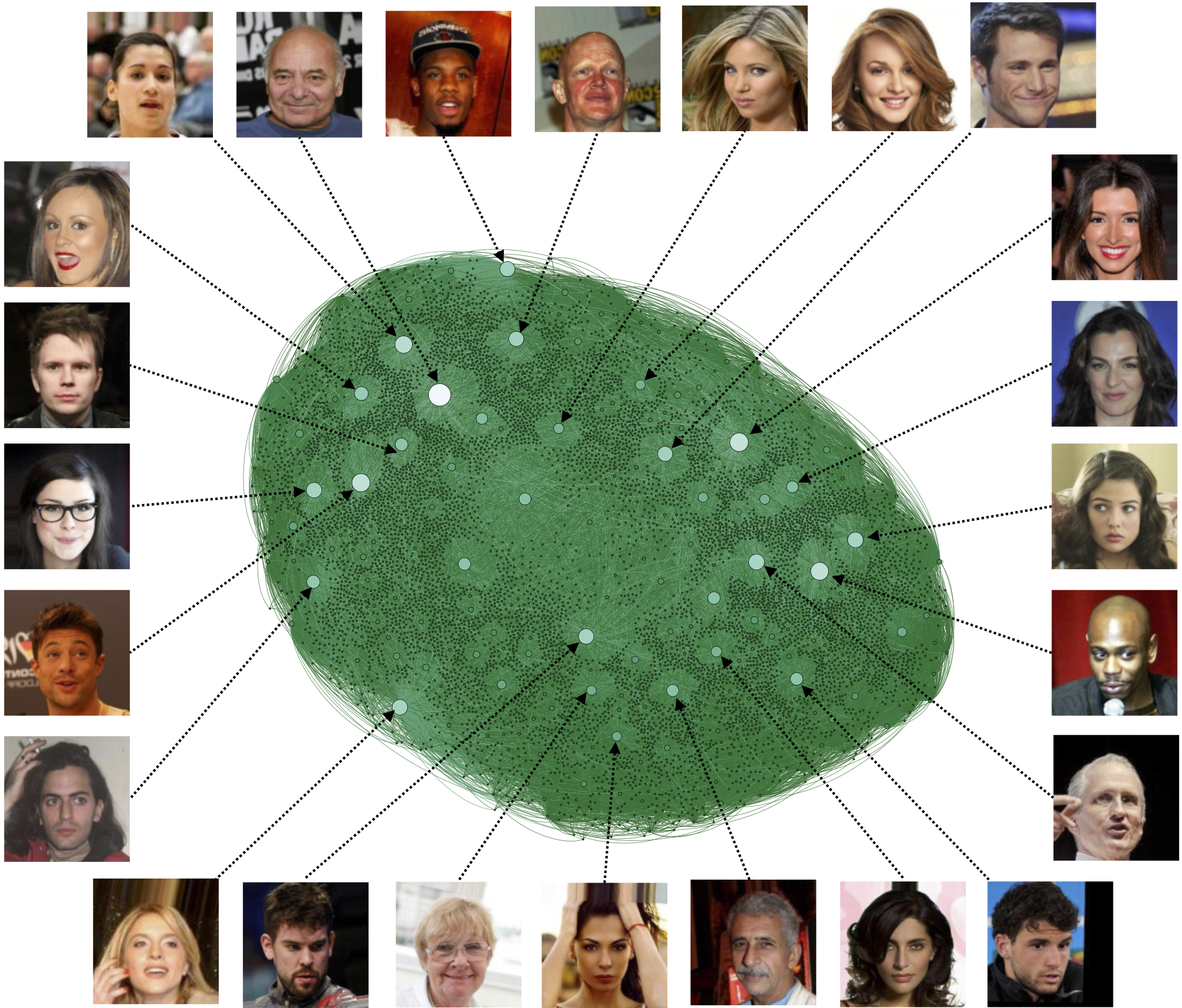}
\caption{Visualization of CelebA identities of adversary classification network. The network (green) corresponds the $k-$nearest neighborhood size $k=5$.}
\label{fig:hub-faces}
\end{figure*}

This section provides a visually more detailed analysis of how the application of focal entropy promotes the formation of  ``hubs''. 
To study that, we analyzed the identity remapping of IDs on the CelebA dataset. Employing focal entropy results in a surjective ID confusion pattern by taking similar IDs into account for privacy sanitization.

\section{Visualization of Hub Faces} 
\label{sec:vishub_face}
To study hubs' semantics, we visualize the CelebA identities of the network corresponding to focal entropy with $k-$nearest neighborhood size $k=5$. See Fig.~\ref{fig:hub-faces} for the visualization of the hub faces. As can be seen, the hubs exhibit a rich diversity in facial properties.

\section{Adversarial Identity Mapping}
\label{sec:vishub_id}
\begin{figure*}[h!]
\centering
\includegraphics[width=1.0\textwidth]{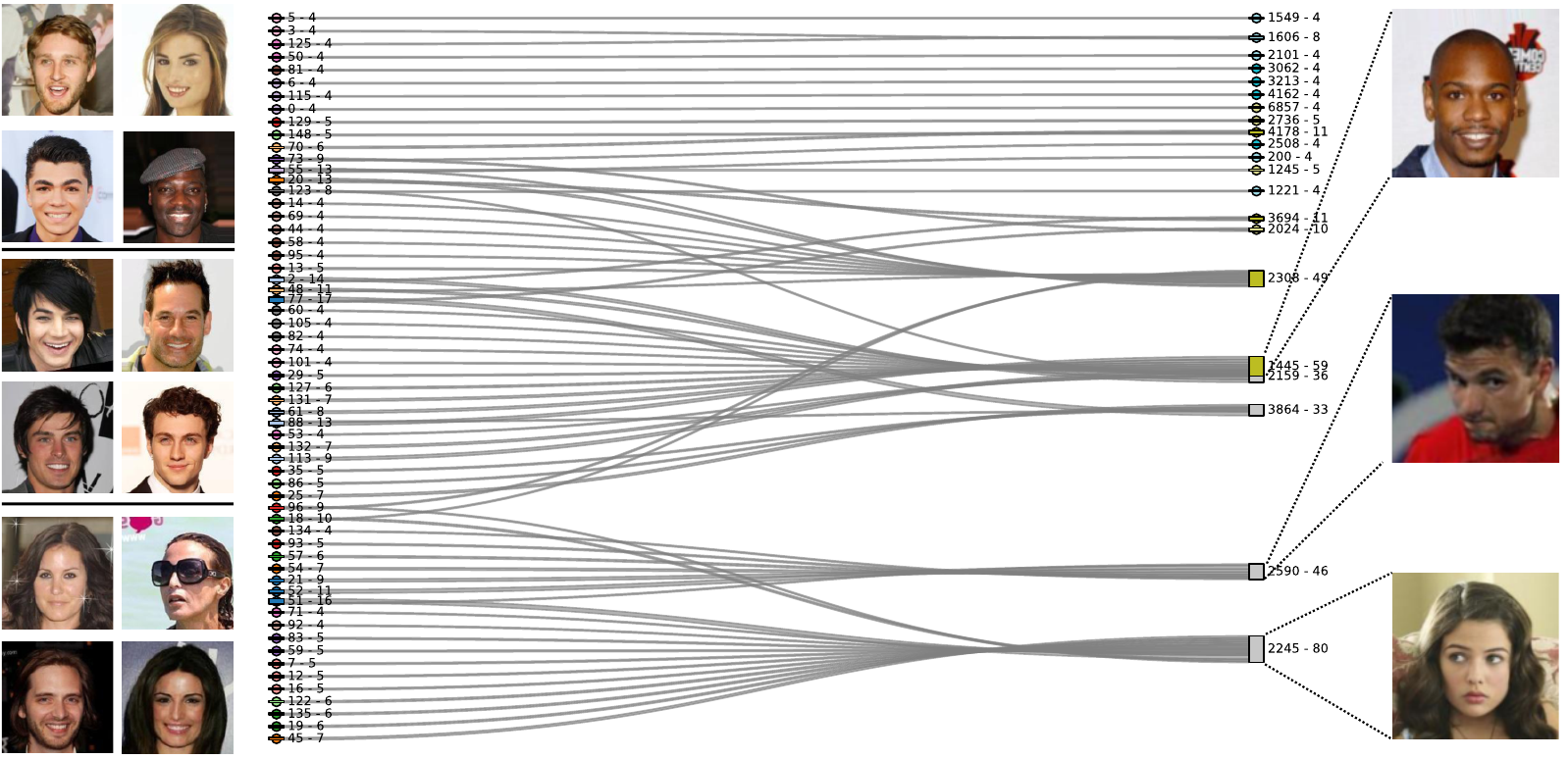}
\caption{Visualization of the remapping of IDs in CelebA due to adversarial representation learning. Source IDs (left) are remapped to new target IDs (right). Pictures on the left are samples that get mapped to a hub; separation with bar indicates different target hub. Pictures on the right are the visualization of hub identities. Visualization contains a subset of 150 IDs, with targets getting at least four associations. The first number at each node indicates the ID, the second is the number of images per ID. Node splicing indicates the remapping of a single ID to multiple adversarial targets.}
\label{fig:network-sankey-faces}
\end{figure*}
Figure~\ref{fig:network-sankey-faces} is a zoom-in version of a graph shown in the main paper, with $k-$nearest neighborhood size $k=5$. This visualization provides a more in-depth view of how the adversarial process leads to a remapping of identities. To avoid visual clutter, a subset of identities and targets was chosen. As can be seen, instead of being a collapse of a facial stereotype, each hub is associated with diverse identities, giving rise to the deep sanitization of the representation.

\section{Attribute-level Analysis}\label{sec:Attribute-level}

We extend the results from the main paper by reporting detailed results of the normalized $\Delta$-Accuracy and privacy trade-off on \emph{all the attributes} in CelebA dataset (Tab. \ref{tab:correlation-analysis}). While in the paper, we only reported the results on the subset of the attributes chosen to be ``relatively unbiased'' in terms of occurrence, here we additionally provide the accuracy on the complete list of attributes.

\begin{table*}[]
\centering
\begin{tabular}{c|ccccc}
\hline
\multicolumn{6}{c}{\textbf{CIFAR-100}~\cite{Krizhevsky09learningmultiple}} \\
\hline
Group &Attribute & Target Accuracy & $\Delta$-Accuracy & Correlation & Prior\\
\hline
\hline
Low & Mouth\_Slightly\_Open & 0.92 & 0.187 & 0.222 & 0.483 \\
Low & Smiling & 0.912 & 0.16 & 0.239 & 0.482 \\
Low & High\_Cheekbones & 0.861 & 0.119 & 0.267 & 0.455 \\
Low & Wavy\_Hair & 0.801 & 0.0248 & 0.326 & 0.32 \\
Low & Oval\_Face & 0.738 & 0.00962 & 0.332 & 0.284 \\ \hdashline
- & Attractive & 0.789 & -0.0144 & 0.335 & 0.513 \\
- & Pointy\_Nose & 0.752 & -0.0248 & 0.351 & 0.277 \\
- & Straight\_Hair & 0.791 & -0.0437 & 0.356 & 0.208 \\
- & Bags\_Under\_Eyes & 0.835 & 0.0151 & 0.358 & 0.205 \\
- & Brown\_Hair & 0.849 & 0.00913 & 0.361 & 0.205 \\
- & Arched\_Eyebrows & 0.826 & -0.006 & 0.369 & 0.267 \\
- & Wearing\_Earrings & 0.865 & 0.0569 & 0.371 & 0.189 \\
- & Big\_Nose & 0.814 & -0.0635 & 0.396 & 0.235 \\
- & Narrow\_Eyes & 0.89 & 0.0404 & 0.402 & 0.115 \\
- & Bangs & 0.949 & 0.0819 & 0.405 & 0.152 \\
- & Bushy\_Eyebrows & 0.893 & -0.0505 & 0.419 & 0.142 \\
- & Blond\_Hair & 0.936 & -0.0159 & 0.429 & 0.148 \\
- & Big\_Lips & 0.78 & -0.0721 & 0.434 & 0.241 \\ \hdashline
Intermediate & Black\_Hair & 0.87 & -0.00142 & 0.370 & 0.239 \\
Intermediate & Heavy\_Makeup & 0.89 & 0.0131 & 0.380 & 0.387 \\
Intermediate & Wearing\_Necklace & 0.882 & 0.00404 & 0.403 & 0.123 \\
Intermediate & Wearing\_Lipstick & 0.906 & -0.00617 & 0.415 & 0.472 \\
Intermediate & No\_Beard & 0.935 & -0.00149 & 0.435 & 0.835 \\ \hdashline
- & 5\_o\_Clock\_Shadow & 0.913 & -0.0467 & 0.435 & 0.111 \\
- & Wearing\_Necktie & 0.941 & 0.0509 & 0.437 & 0.0727 \\
- & Receding\_Hairline & 0.931 & -0.0212 & 0.439 & 0.0798 \\
- & Blurry & 0.956 & 0.125 & 0.443 & 0.0509 \\
- & Rosy\_Cheeks & 0.95 & 0.0581 & 0.448 & 0.0657 \\
- & Eyeglasses & 0.987 & 0.0752 & 0.457 & 0.0651 \\
- & Chubby & 0.952 & -0.0341 & 0.458 & 0.0576 \\
- & Wearing\_Hat & 0.982 & 0.0657 & 0.459 & 0.0485 \\
- & Double\_Chin & 0.958 & -0.0421 & 0.46 & 0.0467 \\
- & Pale\_Skin & 0.967 & 0.108 & 0.461 & 0.0429 \\
- & Goatee & 0.955 & -0.0446 & 0.461 & 0.0628 \\
- & Young & 0.866 & -0.0211 & 0.462 & 0.774 \\ \hdashline
High & Sideburns & 0.957 & -0.0531 & 0.463 & 0.0565 \\
High & Gray\_Hair & 0.971 & -0.0344 & 0.471 & 0.0419 \\
High & Mustache & 0.965 & -0.0413 & 0.474 & 0.0415 \\
High & Bald & 0.984 & -0.0332 & 0.486 & 0.0224 \\
High & Male & 0.958 & -0.0262 & 0.494 & 0.417 \\ \hline
\end{tabular}
\caption{CelebA attribute-level privacy analysis. The group label indicates whether, and for which group attribute was selected for visualization of Fig.~3 in the main paper. 
Accuracy on the target and residual representation part, respectively. Prior is the attribute frequency bias in dataset split. }
\label{tab:correlation-analysis}
\end{table*}

\section{Trade-off Curve}\label{sec:Trade-off}
In this section, we provide a zoom-in version of the trade-off figure in the paper, which shows the results in the in the region of interest, and compares it with several competitors.

\section{Qualitative Results}\label{sec:qual}

Figure~\ref{fig:facerecon} 
shows different reconstructions of additional CelebA identities (equal male and female) at different privacy levels. Each column is two different samples from CelebA (one male and one female), and from top to bottom, the privacy disclosure is decreasing for each. It can be noticed that visualizations from the residual latent part and target latent part confirm the sanitization visually.

\section{Architectural Details}\label{sec:arch}
We describe the architecture of each part of our model. Table~\ref{tab:encoder_decoder} shows the architectures of the VAE, i.e., the encoder and the decoder. It should be noted that the last two layers of the encoder in Tab.~\ref{tab:architecture_encoder} arise from layer splitting to accommodate for partitioning target and residual representations. This is highlighted with dashed lines. Furthermore, we provide the architectures of classifiers in Tab.~\ref{tab:all_classifiers}. Architectures for target and adversarial classifiers are identical.

\begin{figure*}[pb!]
\centering
\includegraphics[width=0.99\textwidth]{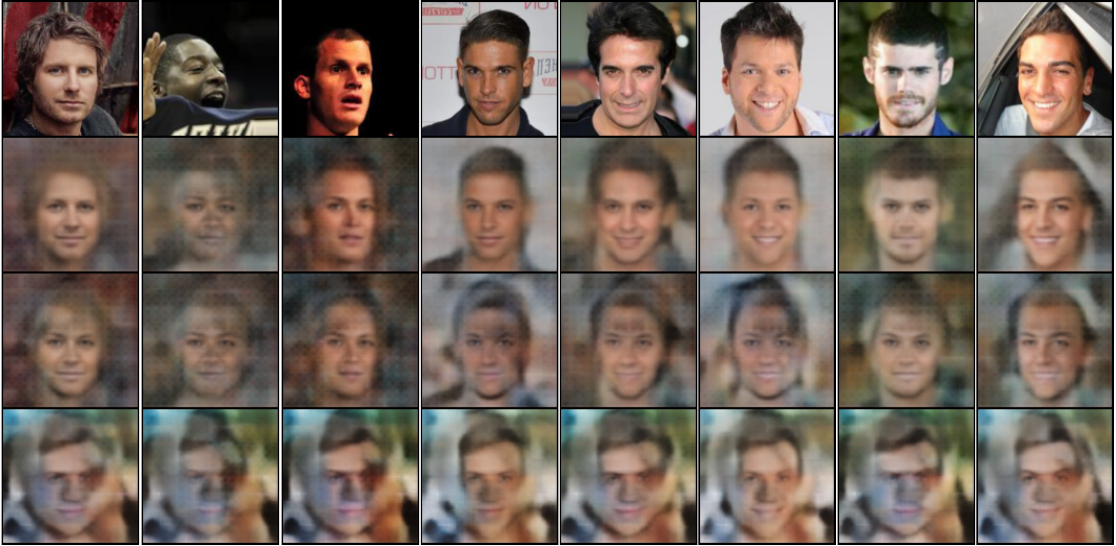}
\vspace{\baselineskip}
\includegraphics[width=0.99\textwidth]{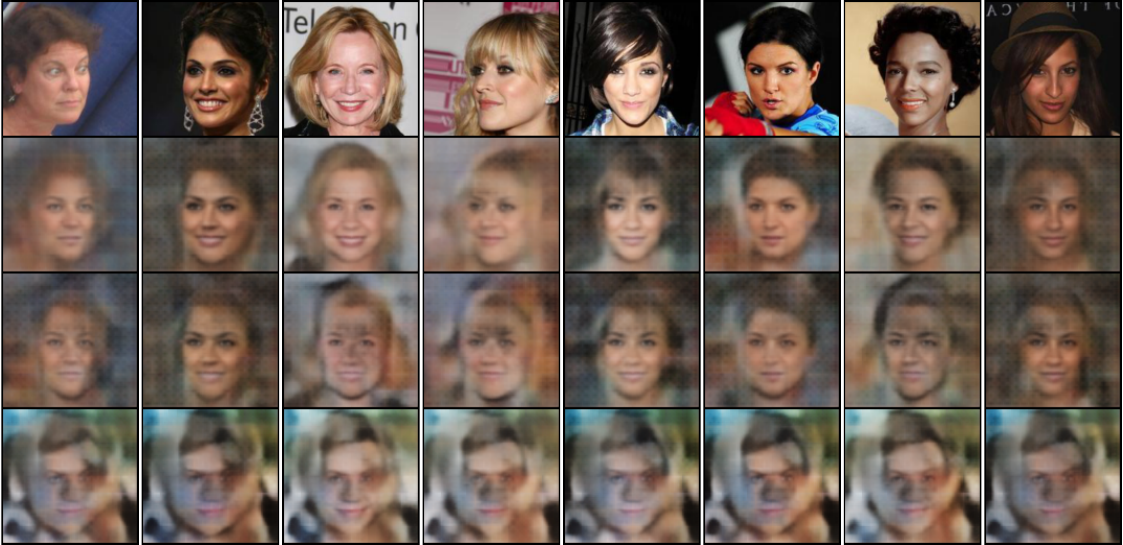}
\caption{Visualization of CelebA data and reconstructions at different privacy levels. From top to bottom, privacy revelation is decreasing. }
\label{fig:facerecon}
\end{figure*}

\newpage

\begin{table*}[h!]
\begin{minipage}{2in}
\centering
\begin{tabular}[t]{l|l|l}
\hline
\textbf{Layer} & \textbf{Output} & \textbf{Parameters}\\
\hline\hline
\multicolumn{3}{c}{Input: $128 \times 128 \times 3$} \\ 
\hline
Conv-2d & $64\times64$ & $64 \times \left[3 \times 3\right]$, st. 2 \\
\hline
\multicolumn{3}{l}{BatchNorm} \\\hline
LeakyReLU & \multicolumn{2}{l}{negative slope: 0.01} \\\hline
Conv-2d & $32\times32$ & $128 \times \left[3 \times 3\right]$, st. 2 \\
\hline
\multicolumn{3}{l}{BatchNorm} \\\hline
LeakyReLU & \multicolumn{2}{l}{negative slope: 0.01} \\\hline
Conv-2d & $16\times16$ & $256 \times \left[3 \times 3\right]$, st. 2 \\
\hline
\multicolumn{3}{l}{BatchNorm} \\\hline
LeakyReLU & \multicolumn{2}{l}{negative slope: 0.01} \\\hline
Conv-2d & $8\times8$ & $512 \times \left[3 \times 3\right]$, st. 2 \\
\hline
\multicolumn{3}{l}{BatchNorm} \\\hline
LeakyReLU & \multicolumn{2}{l}{negative slope: 0.01} \\\hline\hdashline
Linear & \multicolumn{1}{l}{$1\times4096$}& \\ 
Linear & \multicolumn{2}{l}{$1\times512$} \\\hdashline
Linear & \multicolumn{1}{l}{$1\times4096$}& \\ 
Linear & \multicolumn{2}{l}{$1\times512$} \\\hdashline
\hline
\end{tabular}
\centering
\caption{Encoder }
\label{tab:architecture_encoder}
\end{minipage}
\hfillx
\begin{minipage}{2.5in}
\centering
\begin{tabular}[t]{l|l|l}
\hline
\textbf{Layer} & \textbf{Output} & \textbf{Parameters} \\
\hline\hline
\multicolumn{3}{c}{Input: $1024$} \\ 
\hline
Linear & \multicolumn{2}{l}{$32768$} \\\hline
\multicolumn{3}{l}{BatchNorm} \\\hline
LeakyReLU & \multicolumn{2}{l}{negative slope: 0.01} \\\hline
DeConv-2d & $16\times16$ & $256 \times \left[3 \times 3\right]$, st. 2 \\
\hline
\multicolumn{3}{l}{BatchNorm} \\\hline
LeakyReLU & \multicolumn{2}{l}{negative slope: 0.01} \\\hline
Conv-2d & $32\times32$ & $128 \times \left[3 \times 3\right]$, st. 2 \\
\hline
\multicolumn{3}{l}{BatchNorm} \\\hline
LeakyReLU & \multicolumn{2}{l}{negative slope: 0.01} \\\hline
Conv-2d & $64\times64$ & $64 \times \left[3 \times 3\right]$, st. 2 \\
\hline
\multicolumn{3}{l}{BatchNorm} \\\hline
LeakyReLU & \multicolumn{2}{l}{negative slope: 0.01} \\\hline
Conv-2d & $128\times128$ & $3 \times \left[3 \times 3\right]$, st. 2 \\
\hline
\multicolumn{3}{l}{Tanh} \\\hline
\end{tabular}
\centering
\label{tab:architecture_decoder}
\caption{Decoder }
\end{minipage}
\caption{Architectural details of VAE components. Parameters for convolutions correspond to: number kernels $\times [$ kernel size $]$, and stride. Dashed lines in the encoder denote the two separate streams.}
\label{tab:encoder_decoder}
\end{table*}

\begin{table*}
\centering
\begin{subtable}{.48\textwidth}
\centering
\begin{tabular}{l|l}
\hline
\textbf{Layer} & \textbf{Output size / Params} \\
\hline\hline
Linear & \multicolumn{1}{l}{256} \\\hline
\multicolumn{2}{l}{BatchNorm} \\\hline
\multicolumn{2}{l}{PReLU} \\\hline
Dropout & \multicolumn{1}{l}{drop-rate: 0.2} \\\hline
Linear & \multicolumn{1}{l}{128} \\\hline
\multicolumn{2}{l}{BatchNorm} \\\hline
\multicolumn{2}{l}{PReLU} \\\hline
Linear & \#classes \\\hline
\end{tabular}
\caption{Classifier on CIFAR-100}
\label{tab:architecture_cifar_classifier}
\end{subtable}
\begin{subtable}{.48\textwidth}
\centering
\begin{tabular}{l|l}
\hline
\textbf{Layer} & \textbf{Output size / Params} \\
\hline\hline
Linear & \multicolumn{1}{l}{256} \\\hline
\multicolumn{2}{l}{BatchNorm} \\\hline
\multicolumn{2}{l}{PReLU} \\\hline
Dropout & \multicolumn{1}{l}{drop-rate: 0.5} \\\hline
Linear & \multicolumn{1}{l}{128} \\\hline
\multicolumn{2}{l}{BatchNorm} \\\hline
\multicolumn{2}{l}{PReLU} \\\hline
Linear & \#IDs or \#attributes $\times \left[1\right]$  \\\hline
\end{tabular}
\caption{Normal classifier on CelebA}
\label{tab:architecture_celeba_id_classifier}
\end{subtable}
\vspace{\baselineskip}
\begin{subtable}{.48\textwidth}
\centering
\begin{tabular}{l|l}
\hline
\textbf{Layer} & \textbf{Output size / Params} \\
\hline\hline
Linear & \multicolumn{1}{l}{256} \\\hline
\multicolumn{2}{l}{BatchNorm} \\\hline
\multicolumn{2}{l}{PReLU} \\\hline
Dropout & \multicolumn{1}{l}{drop-rate: 0.5} \\\hline
Linear & \multicolumn{1}{l}{128} \\\hline
\multicolumn{2}{l}{BatchNorm} \\\hline
\multicolumn{2}{l}{PReLU} \\\hline
Dropout & \multicolumn{1}{l}{drop-rate: 0.5} \\\hline
Linear & \multicolumn{1}{l}{256} \\\hline
\multicolumn{2}{l}{BatchNorm} \\\hline
\multicolumn{2}{l}{PReLU} \\\hline
Linear & \#IDs \\\hline
\end{tabular}
\caption{Strong ID classifier on CelebA}
\label{tab:strong_id_classifiers}
\end{subtable}
\caption{Architectural details of the used classifiers}
\label{tab:all_classifiers}
\end{table*}

\end{document}